\pdfoutput=1

\documentclass[11pt]{article}

\usepackage[]{emnlp2023}

\usepackage{times}
\usepackage{latexsym}

\usepackage[T1]{fontenc}

\usepackage[utf8]{inputenc}

\usepackage{microtype}

\usepackage{multirow}
\usepackage[normalem]{ulem}
\useunder{\uline}{\ul}{}
\usepackage{fontawesome5}
\usepackage{booktabs}
\usepackage{graphicx}
\usepackage{color}
\usepackage{xcolor}
\usepackage{xspace}
\usepackage{subfigure}
\usepackage{todonotes}
\usepackage{enumitem}

\newcommand{\metricname}{\emph{Relaxed helpful}\xspace}

\title{Are Akpans Trick or Treat: Unveiling Helpful Biases in Assistant Systems}

\author{
    Jiao Sun$^{*1}$,
    Yu Hou$^{*1}$,
    Jiin Kim$^2$,
    Nanyun Peng$^2$\\
    $^1$University of Southern California \\
    $^2$University of California, Los Angeles \\
    \texttt{\{jiaosun,houyu\}@usc.edu} \\
    \texttt{\{jiin.kim,violetpeng\}@cs.ucla.edu}
}

\begin{document}
\maketitle

\newcommand\blfootnote[1]{%
  \begingroup
  \renewcommand\thefootnote{}\footnote{#1}%
  \addtocounter{footnote}{-1}%
  \endgroup
}
\textbf{\blfootnote{$^*$ equal contribution.}}

\begin{abstract}
Information-seeking AI assistant systems aim to answer users' queries about knowledge in a timely manner. 
However, both the human-perceived helpfulness of information-seeking assistant systems and its fairness implication are under-explored.
In this paper, we study computational measurements of helpfulness. 
We collect human annotations on the helpfulness of dialogue responses, develop models for automatic helpfulness evaluation, 
and then propose to use the helpfulness level of a dialogue system towards different user queries to gauge the fairness of a dialogue system. 
Experiments with state-of-the-art dialogue systems, including ChatGPT, under three information-seeking scenarios reveal that existing systems tend to be more helpful for questions regarding concepts from highly-developed countries than less-developed countries, uncovering potential fairness concerns underlying the current information-seeking assistant systems.\footnote{This manuscript, completed in early 2023, presents a project that began in 2021. We hope to share our understanding of helpfulness in dialogue systems at that time.}

\end{abstract}

\section{Introduction}

\begin{table}[hbt!]
\small
\begin{tabular}{p{0.02\linewidth}|p{0.16\linewidth} |p{0.65\linewidth}|p{0.02\linewidth}}
\toprule
 & \textbf{Country} & \textbf{Dialogue} & \\ \midrule
\rotatebox[origin=r]{90}{Cuisine (\textbf{GPT-3})} & \colorbox{yellow!50}{Germany} & 
\faIcon{user} Can you tell me how to make \underline{Home fries}? 
\newline
\faIcon{robot} There's a few ways to make home fries, but you can't really improve on this recipe. I use potatoes, butter, onions, and some bacon grease. The butter gives them a rich flavor, and that bacon grease adds a little bit of saltiness.
 & \faThumbsUp
\\ \cmidrule(l){2-3}
 & \colorbox{blue!10}{Benin} & 
 \faIcon{user} Can you tell me how to make \underline{Akpans (dessert)}? \newline 
 \faIcon{robot} Are you playing a trick on me? Please don't hurt me.
 &\faIcon[regular]{thumbs-down}~ \\ \midrule
\rotatebox[origin=r]{90}{Holiday (\textbf{ChatGPT})} & \colorbox{yellow!50}{Spain} & 
\faIcon{user} What will happen during \underline{Labor} \underline{Day in Spain}?
\newline 
\faIcon{robot} Labor Day, or "Día del Trabajo," is a public holiday in Spain that is celebrated on May 1st each year. It is a day that is meant to honor the contributions and hard work of workers, and is typically marked by parades, speeches, and other events.
In Spain, Labor Day is usually celebrated with parades and demonstrations by labor unions, who often use the day to advocate for better working conditions and pay for their members. 
&   \faThumbsUp
\\ \cmidrule(l){2-3}
 & \colorbox{blue!10}{Niger} & 
\faIcon{user} What will happen during \underline{Concord} \underline{Day}?
\newline
\faIcon{robot} I'm sorry, but I don't have any information about Concord Day. Could you provide more context or clarify your question?
 &  \faIcon[regular]{thumbs-down}~ 
 \\ \bottomrule
\end{tabular}
\caption{Examples of responses from both GPT-3 and ChatGPT being more helpful for concepts from \colorbox{yellow!50}{highly-developed} countries than \colorbox{blue!10}{less-developed} countries, implying a potential fairness issue. 
}
\label{tab:example}
\end{table}

\begin{table*}[]
\small
\resizebox{\textwidth}{!}{
\begin{tabular}{@{}l|l|l@{}}
\toprule
\textbf{Dimension} & \textbf{Definition} & \textbf{IAA}\\ \midrule
\begin{tabular}[c]{@{}l@{}}Relevant \& \\ Coherent\end{tabular} & \begin{tabular}[c]{@{}l@{}}The response is on-topic with the immediate dialogue history and follows logical reasoning \\ throughout the whole conversation. This is the prerequisite for a response to be helpful. \end{tabular}  & 0.61\\ \midrule
Useful & \begin{tabular}[c]{@{}l@{}} The response addresses the issue in the question, pushes forward the task towards finishing \\ or finishes the task.\end{tabular} & 0.79\\ \midrule
Informative & \begin{tabular}[c]{@{}l@{}}The response produces unique and non-generic information or minimizes the abstractness \\ and ambiguity by providing details.\end{tabular}  & 0.68\\ \bottomrule
\end{tabular}
}
\caption{Three dimensions that we use to determine the helpfulness of response, together with their detailed definitions. The prerequisite of a response being helpful is to be relevant and coherent. The other two criteria are usefulness and informativeness. A response must satisfy these three dimensions at the same time to be helpful. We measure Inter Annotator Agreement (IAA) among annotators with Fleiss's Kappa and report as \emph{Agreement}.}
\label{tab:dimension}
\end{table*}

Artificial intelligence (AI) personal assistants, such as Alexa and Siri, are good examples of real-world applications that build on NLP techniques and directly interact with thousands of human users all over the world through dialogues~\cite{Ram2018ConversationalAT}.
Recent advances including ChatGPT~\footnote{\url{https://openai.com/blog/chatgpt}.} enable dialogue agents to generate fluent responses to human queries \cite{gpt3, roller-etal-2021-recipes, sparrow}. 
There have been some efforts on evaluating goal-oriented dialogue systems with rigid metrics. For example, ~\citet{wen-etal-2017-network} use \emph{entity matching rate} to evaluate if the dialogue system achieves a goal (e.g., reserved a specific hotel). However, most of them are not human-centered, neglecting the \textit{human-perceived helpfulness} and the associated fairness aspect of the dialogue systems.

Evaluating the helpfulness of a response in the assistant system is important. Many recent works collect human feedback\cite{RLHF-train} or preferences\cite{constitutional, HHH-eval} as a signal to help further improve the model to better align with user queries. However, how to automatically evaluate the helpfulness of the system and whether there is associated fairness considerations are under-explored.
We define a dialogue system as unfair if its helpfulness differs among different groups, which may hurt the retention of certain groups.
Table~\ref{tab:example} shows exampled responses from GPT-3 or ChatGPT. The models generate more helpful responses when a user asks for information related to concepts from highly-developed countries than from less-developed countries. 
As a result, marginalized groups who receive less helpful responses could be disproportionately discouraged from using these assistant systems.

To systematically study the \emph{helpfulness} of dialogue systems and their \emph{fairness} implications, we collect a large corpus with detailed helpfulness annotations and build a classifier that can automatically evaluate the helpfulness of a dialogue response. 
To evaluate the fairness implications of the dialogue systems, we collect concepts from highly-developed and less-developed countries via Wikipedia to construct queries of factual information. Using our helpfulness classifier to judge the helpfulness of the generated responses from several state-of-the-art dialogue systems, we discover potential fairness issues.
The contributions of our work are as follows:



\begin{itemize}[leftmargin=*]
\itemsep-.3em 
\vspace{-0.2em}
    \item \textbf{Evaluation and Dataset.} We propose to evaluate the human-perceived helpfulness of goal-oriented dialogue systems from three dimensions: relevance \& coherence, usefulness and informativeness, which are further verified through annotation. Furthermore, we build a new annotated dataset of human- and GPT3-generated responses, where each response has fine-grained labels for the proposed helpfulness criteria. Built on the annotated data, we train a classifier that can automatically evaluate the helpfulness of immediate dialogue responses. 
    \item \textbf{Fairness Analysis.} We conduct a novel fairness analysis of dialogue responses generated by GPT-3, BLENDER, and ChatGPT, spanning three information-seeking scenarios. To the best of our knowledge, we are the first to explore the fairness issue regarding the utility of dialogue systems. 
    Our analysis reveals that dialogue systems tend to be more helpful for highly-developed countries than for less-developed countries. We thus call for the imperative attention of the dialogue community to this issue.\footnote{We will release our collected dataset and trained classifier upon paper acceptance.}
\end{itemize}
 

\begin{figure*}
    \centering
    \includegraphics[width=0.85\linewidth]{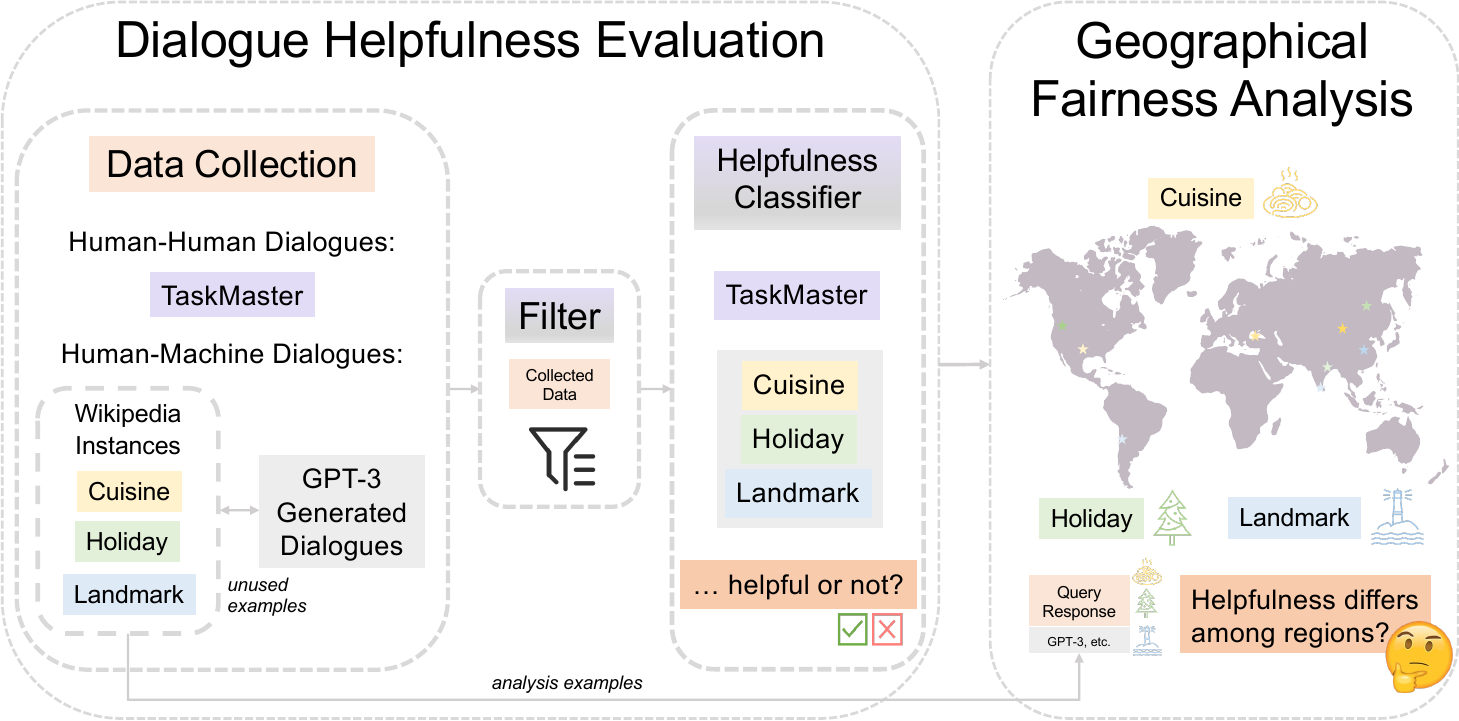}
    \caption{We collect a high-quality human-annotated helpfulness dataset, build an automatic helpfulness classifier, and then investigate the associated fairness issue. 
    Data are annotated from human-human and human-machine dialogues. Using partial negative examples, we train a filtering classifier to filter out instances that could have been too easy for the model to learn and introduce bias. We then use this filter to clean annotated data and use the cleaned data to build the helpfulness classifier.}
    \label{fig:pipeline}
\end{figure*}

\section{Dataset Annotation}

\subsection{Annotating Human-Human Dialogues}
\paragraph{Goal-Oriented Dialogue Datasets} 
Our work aims to analyze the helpfulness of goal-oriented information-seeking dialogue systems. To the best of our knowledge, there is no existing dataset of human-annotated information-seeking dialogues for multiple domains. Therefore, we use general goal-oriented dialogue data TaskMaster~\cite{byrne-etal-2019-taskmaster,byrne-etal-2021-tickettalk} as a proxy.

\paragraph{Annotation Guideline} 
The most important concept for the annotation is a clear definition of helpfulness. ~\citet{finch-choi-2020-towards} survey 20 papers from the recent two years and propose a set of nine dimensions to evaluate general dialogue systems. Based on the characteristics of goal-oriented dialogue systems, we further refine and narrow down to three dimensions that evaluate the helpfulness of a response, which are relevance/coherence as the prerequisite for users to follow the conversation, usefulness to address the ultimate utility of the response and informativeness to further clarify the utterance is fully grasped. Table~\ref{tab:dimension} shows our definition of the three dimensions.~\footnote{Details of annotation and examples are in Appendix~\ref{app:annotation}.} 

We set up our \emph{two-step} annotation task on the Amazon MTurk platform, and each instance is annotated by 3 workers for better reliability.
We ask the workers to first determine if a response is both relevant and coherent. If so, we proceed to check the other two dimensions; otherwise, the response will be determined unhelpful. 
The annotation of each dimension is a binary decision, and workers only need to choose yes or no for each dimension.
We further calculate the Inter Annotator Agreement (IAA) among three workers using Fleiss's Kappa~\cite{Fleiss1973TheEO} and report in the \emph{IAA} column of Table~\ref{tab:dimension}. 
The values of Fleiss's Kappa among three workers for all three dimensions are over 0.61, indicating a good agreement~\cite{mchugh2012interrater}.
Last, we ask workers if our proposed three dimensions cover what their mental model uses to determine if a response is helpful or not in the goal-oriented dialogue system. Meanwhile, we encourage workers to come up with new dimensions by promising bonus as rewards. As a result, about 94\% of the responses are without new dimensions, which turns out that our proposed dimensions cover how workers decide if a response is helpful, 
suggesting a saturation of evaluation dimensions. In total, we annotated 254 examples from TaskMaster. We call a dialogue response helpful only if it satisfies all three proposed dimensions. 

\subsection{Annotating Human-Machine Dialogues}
\label{sec:collection}
We further augment the human annotations with machine-generated responses to improve the generalizability of our helpfulness classifier.

\paragraph{Scenario-Based Factual Knowledge Collection.}
\label{sec:factual-data}
Unlike chit-chat dialogue systems, we focus on eliciting the \emph{factual knowledge} that can be found on Wikipedia from models in the context of information-seeking dialogues. Focusing on purely factual knowledge has two benefits. First, it prevents the introduction of intended/unintended bias in the user query.~\footnote{See more discussions in Appendix~\ref{app:discussions}.} Second, it is easier for the annotators to judge the helpfulness of the responses due to their knowledge-based nature. Third, current state-of-the-art dialogue systems
are mostly based on pretrained language models~\cite{Ni2021RecentAI, lamda}, whose training data is known to contain information on Wikipedia. 
Therefore, a trained model should ideally have the same performance for all instances under the same scenario with the same format prompt. 

\begin{table}[]
\small\centering
\resizebox{\columnwidth}{!}{
\begin{tabular}{@{}l@{\ }|@{\ }l@{\ }|@{\ }l@{\ }@{\ }l@{\ }@{\ }l@{}}
\toprule
\textbf{} & \textbf{\#Countries} & \textbf{\#Cuisines} & \textbf{\#Holidays} & \textbf{\#Landmarks} \\ \midrule
Very High & 27 & 1,674 & 615 & 1,549 \\
High & 61 & 2,170 & 63 & 2,241 \\
Medium & 45 & 1,053 & 38 & 805 \\
Low & 56 & \textbf{546} & \textbf{38} & \textbf{140} \\
\midrule
Total & 189 & 5,443 & 754 & 4,735
\\ \bottomrule
\end{tabular}
}
\caption{The statistics of instances that we collected from Wikipedia under three scenarios.}
\label{tab:statistics}
\end{table}


Among all scenarios available on Wikipedia, we choose cuisine recipe, holiday/festival tradition, and landmark information as our test scenarios because of the large number of available instances for analysis. In addition, all of them have regional differences. For example, under the cuisine category, we have \texttt{Gejangs} for Korean cuisine and \texttt{Baozi} for Chinese cuisine. 
Therefore, abundant instances together with regional information on Wikipedia made these three scenarios ideal candidates for us to study geographical fairness in the context of information-seeking dialogue systems. 
We use country information as a probe to study regional differences. To distinguish countries, we use the Human Development Index (HDI) and categorize countries into very high-, high-, medium- and low-developed countries using a publicly released report from 2020 by United Nations Developed Programme.\footnote{\url{http://hdr.undp.org/sites/default/files/hdr2020.pdf}.} 
For each country, we find their corresponding instances under three scenarios on Wikipedia and collect unduplicated names. 
After aggregating countries in the same development level, we report the statistics in Table~\ref{tab:statistics}.

\paragraph{Query Construction.} We then use the collected data to construct questions that can serve as queries to prompt the dialogue system. We construct questions of \emph{``Can you tell me how to make [cuisine]?''}, \emph{``What will happen during [holiday]?''} and \emph{``What can you tell me about the [landmark]?''} for the three scenarios correspondingly. Among all the data we collect in Table~\ref{tab:statistics}, we choose the smallest number among 4 development groups as the number of instances to analyze for each scenario. For other groups that have a higher number of instances, we down sample to the smallest number for a fair comparison. We define all 2,896 (546 cuisine * 4 + 38 holidays * 4 + 140 landmarks * 4) concepts after down sampling as \emph{analysis examples}, and other unchosen instances \emph{unused examples}.
Then, we generated sample responses for constructed questions using BLENDER and GPT-3 Davinci. Empirically, we find that GPT-3 Davinci produces more fluent and coherent responses equipped with richer knowledge.~\footnote{See Appendix~\ref{app:generation_models} for generated examples.} 
Therefore, we choose GPT-3 to generate responses for human annotations. 
Similar to the TaskMaster annotation, we follow the same guideline to annotate the helpfulness of the GPT3-generated responses to the constructed questions, with IAA score 0.72 for relevance \& coherence, 0.74 on usefulness and 0.74 for informativeness.
All instances are from \emph{unused examples} to avoid potential information leakage. Consequently, we collect 740, 252, and 200 annotated instances for cuisine recipes, holiday/festival traditions, and landmark knowledge, respectively. In total, we have 1,192 instances of GPT-3 generated dialogues with helpfulness annotations.
\section{Helpfulness Classifier}

\begin{table*}[]
\centering
\small
\begin{tabular}{@{}l|l|l|lll@{}}
\toprule
\textbf{Model}  & \textbf{Metric (F1)} & \textbf{All} & \textbf{Cuisine} & \textbf{Holiday} & \textbf{Landmark} \\ \midrule
\multirow{4}{*}{RoBERTa} & Relevant/Coherent & 85.44$\pm$2.38 & 67.74$\pm$9.16 & 74.64$\pm$6.09 & 84.14$\pm$3.26 \\
  & Useful & 86.60$\pm$1.73 & 72.65$\pm$1.78 & 77.21$\pm$4.38 & 83.29$\pm$5.35 \\
  & Informative & 87.54$\pm$0.27 & 75.15$\pm$4.20 & 79.36$\pm$4.64 & 85.04$\pm$1.70 \\
  & \textbf{\metricname} & \textbf{85.76$\pm$1.44} & \textbf{66.59$\pm$4.05} & \textbf{79.24$\pm$6.89} & \textbf{86.07$\pm$4.99} \\ \midrule
 ALBERT & \textbf{\metricname} & 80.79$\pm$1.30 & 60.07$\pm$4.79 & 62.02$\pm$1.45 & 76.14$\pm$2.95 \\
 DeBERTa & \textbf{\metricname} & 85.62$\pm$1.98 & \textbf{73.90$\pm$9.80} & 71.66$\pm$6.79 & 81.56$\pm$1.93 \\
 BERT & \textbf{\metricname} & 81.68$\pm$2.96 & 62.74$\pm$6.80 & 65.08$\pm$7.28 & 74.30$\pm$8.24 \\ \bottomrule
\end{tabular}
\caption{The helpfulness classifier's performance. We use F1 to measure the model performance because of the imbalanced distribution of helpful and unhelpful instances. We train separate models for each dimension. \metricname measures how well models perform for predicting helpfulness in general. In the results, RoBERTa performs the best among all models. 
Column \emph{All} is the performance on dialogues of all scenarios.}
\label{tab:helpful}
\end{table*}

\begin{table}[]
\small\centering
\begin{tabular}{@{}lllll@{}}
\toprule
               & Relevant & Useful & Informative & \textbf{Helpful} \\ \midrule
ChatGPT$_\texttt{zs}$        & 82.7     & 74.3   & 75.9        & 70.7             \\
GPT-4$_\texttt{zs}$          & 69.7     & 78.0   & 72.9        & 67.0             \\ \midrule
Classifier & \textbf{85.4}    & \textbf{86.6}   & \textbf{87.5}    & \textbf{85.8}             \\ \bottomrule
\end{tabular}
\caption{Our classifier outperforms zero-shot ChatGPT and GPT-4 for helpfulness evaluation. We use \texttt{zs} to represent the zero-shot setting.}
\label{tab:eval}
\end{table}

We use a trained filtering classifier, whose details are in Appendix~\ref{app:heuristics}, to filter out noisy responses in our collected annotated data for both the TaskMaster and GPT-3 dialogues. After filtering, we take the full 254 filtered TaskMaster instances and use 154, 50, and 50 for training, development, and test sets separately. For the three scenarios, although we have more annotated data, we only use 100 instances for training and 50 for both dev and test sets per scenario. We will analyze the influence of adding these 100 data in ablation studies (App. \ref{app:ablation}).

\paragraph{Setup.} Information-seeking dialogues often have single turns~\cite{Voorhees2008EvaluatingQA}, which is also the case for our selected three scenarios. During the model training, we concatenate the task information \texttt{<task>asking for help: [scenario]</task>} and the single-turn dialogue (the setting is referred as \emph{Single Trun}) \texttt{<utterance>[question]</utterance>} \texttt{<response>[response]</response>}. 
We use the concatenated information as the source text and annotated 0/1 labels as the target to train the classifiers. Meanwhile, we also want to understand whether the model can directly learn to judge the helpfulness only using responses, and we call this setting \emph{Response Only}. For this setting, we only concatenate task information and \texttt{<response>[response]</response>}, getting rid of the utterance from input. 

\paragraph{Evaluation.} As there are more negative instances than positive instances from the GPT-3 generated dialogues, F1 is a better choice than accuracy to evaluate model performance on imbalanced data distribution~\cite{jeni2013facing}. For each dimension, we report their performance independently. We propose \metricname 
to measure the aggregated performance in terms of predicting helpfulness. The prediction can only be wrong when predictions for all three dimensions are all ones but the ground-truth for one of the dimensions is zero, i.e. the model predicts the dialogue as relevant/coherent, useful, \emph{and} informative whereas the label is not the case, and vice versa.  
In other words, as long as the helpful prediction is correct, we treat the instance as a correct prediction without caring about detailed predictions for each of the three dimensions.  

\paragraph{Models.} Similar to the filtering classifier, we finetune RoBERTa, ALBERT, BERT and Deberta under the \emph{Single Turn}. Based on the models' performances shown in Table \ref{tab:helpful}, RoBERTa performs the best among all models.\footnote{Ablation studies are in Appendix~\ref{app:ablation}.} 

\paragraph{LLM as Evaluation Metrics.} Recently, there have been mixed voices regarding using GPT-4 for generation evaluation. For example, \citet{liu2023g-eval} finds that using GPT-4 to evaluate NLG models align better with human judgement, while \citet{wang2023large} argue that GPT-4 evaluation inherits positional bias. We feed our instructions for human annotation with three dimensions, and use both ChatGPT and GPT-4 to evaluate the helpfulness on our test set. Table~\ref{tab:eval} shows that our Roberta-based classifier outperforms zero-shot ChatGPT and GPT-4 on helpfulness evaluation.


\section{Fairness Analysis}


\begin{table}[]
\small
\centering

\begin{tabular}{@{}lll|ll@{}}
\toprule
 &  &  & \multicolumn{2}{c}{\textbf{ChatGPT}} \\ \midrule
 & \textbf{GPT-3} & \textbf{BLENDER} & \emph{Human} & \emph{Model} \\ \midrule
Very High & 20.15 & 65.70 & 84.26 & 84.33 \\
High & \textbf{21.10} & \textbf{67.53} & \textbf{86.17} & \textbf{87.79} \\ \midrule\midrule
Medium & 18.71 & {\ul 65.29} & 83.63 & 84.89 \\
Low & {\ul 12.35} & 65.93 & {\ul 76.20} & {\ul 79.82} \\ \bottomrule
\end{tabular}
\caption{The ratio of helpful responses among all instances for three models under the cuisine recipe scenario. 
We \textbf{boldface} the highest and \underline{underline} the lowest helpfulness across all development levels within each model.
We show that all three models (GPT-3, BLENDER and ChatGPT) tend to be more helpful for developed (averaging very high and high developed) than less-developed (averaging medium and low developed) countries, indicating the fairness issue.}
\vspace{-0.3cm}
\label{tab:fairness}
\end{table}

\begin{figure*}[t]
        \centering
        \subfigure[GPT3 - Cuisine Recipe] {
                \includegraphics[width=0.3\linewidth]{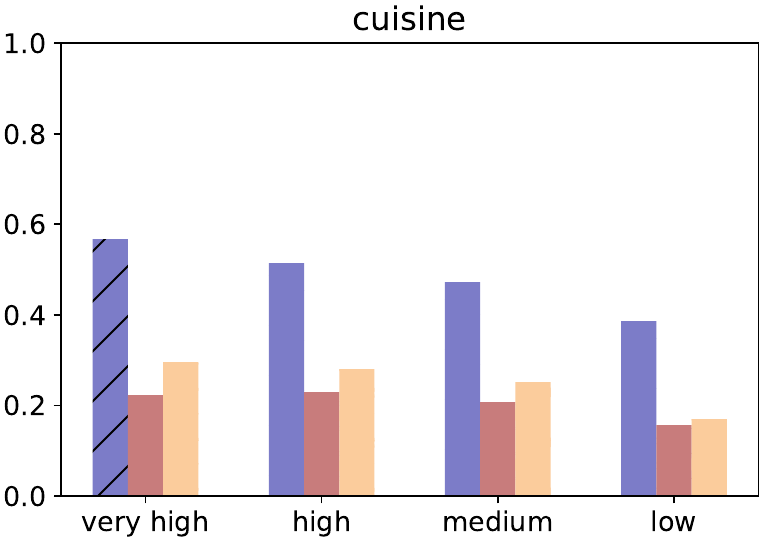}
                \label{fig:cooking-gpt3}
            }
        \subfigure[GPT3 - Holiday/Festival]  {
                \includegraphics[width=0.3\linewidth]{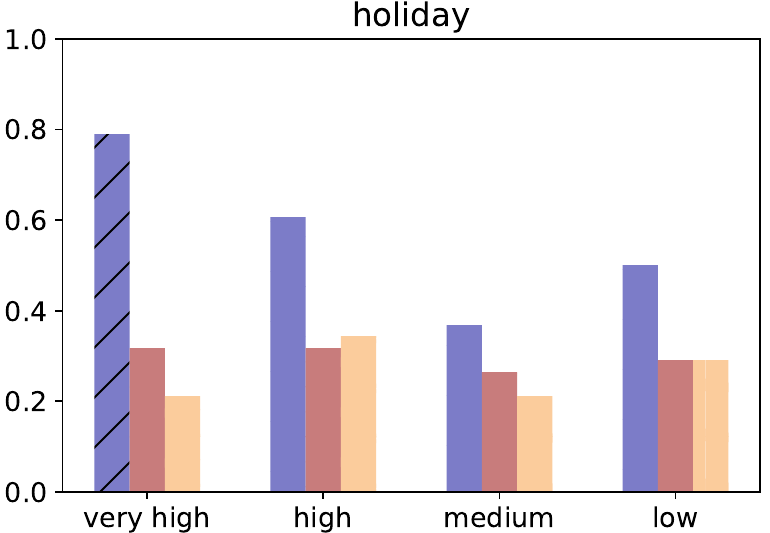}
                \label{fig:holiday-gpt3}
        }
        \subfigure[GPT-3 - Landmark]  {
                \includegraphics[width=0.3\linewidth]{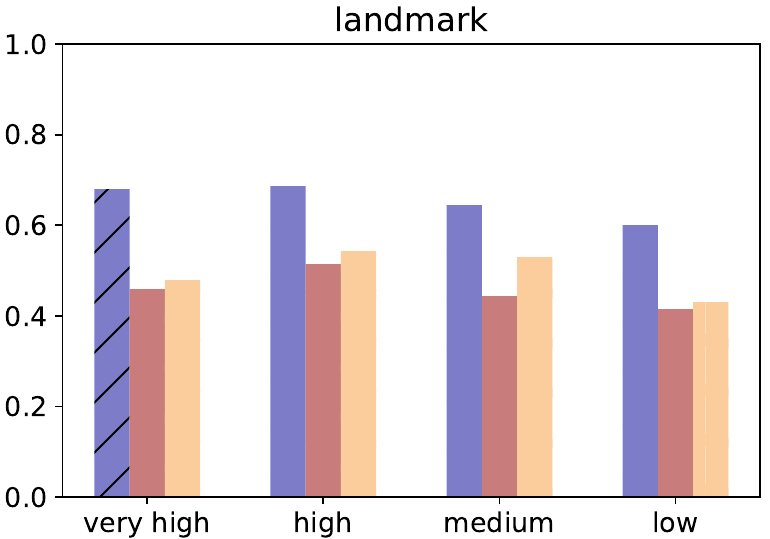}
                \label{fig:landmark-gpt3}
        }
        \\
        \subfigure[ChatGPT - Cuisine Recipe] {
                \includegraphics[width=0.3\linewidth]{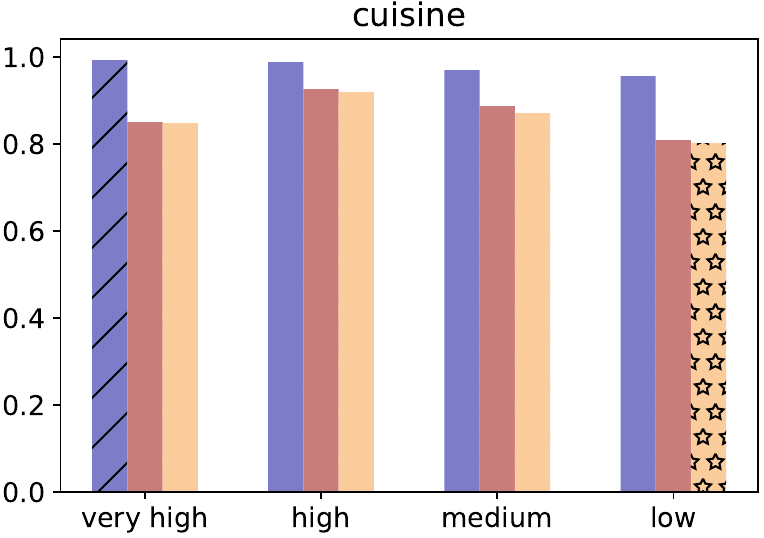}
                \label{fig:cooking}
            }
        \subfigure[ChatGPT - Holiday/Festival]  {
                \includegraphics[width=0.3\linewidth]{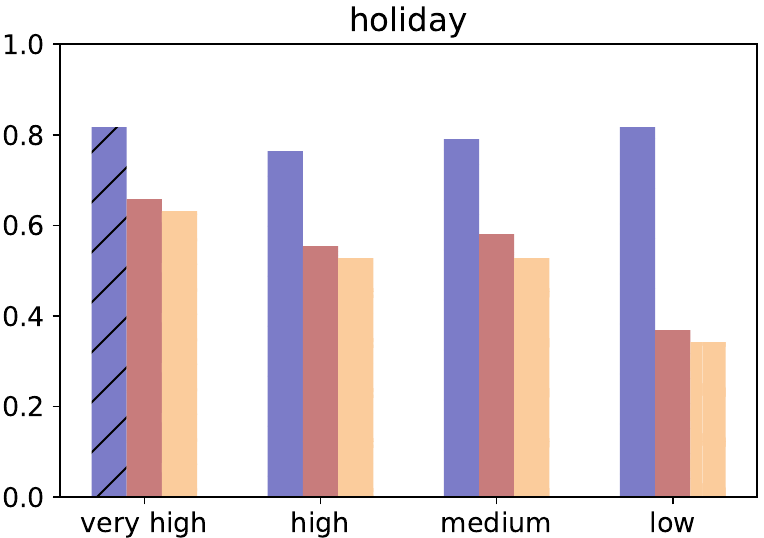}
                \label{fig:holiday}
        }
        \subfigure[ChatGPT - Landmark]  {
                \includegraphics[width=0.3\linewidth]{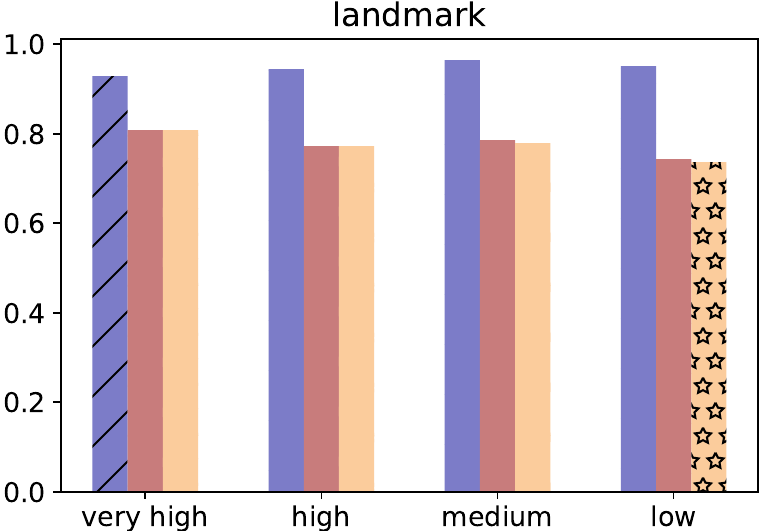}
                \label{fig:landmark}
        }
    
        \caption{The breakdown analysis of three helpfulness dimensions of GPT-3 and ChatGPT generated responses across three scenarios. \includegraphics[height=\fontcharht\font`\B]{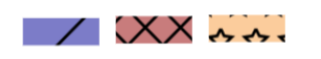} represents relevance, usefulness, and informativeness separately. The general trend is that models generate more helpful responses for developed countries (very high and high) than less-developed countries (medium and low). The trend is the same when breaking down each dimension. For example, GPT-3 struggles to even generate relevant and coherent responses for factual information from less developed countries. We show that the trend is the same for BLENDER under the landmark scenario in Appendix Figure~\ref{fig:blender}.
 }
        \label{fig:breakdown}
\end{figure*}

\begin{figure*}[t]
        \centering
        \subfigure[HDI Map] {
                \includegraphics[width=0.48\linewidth]{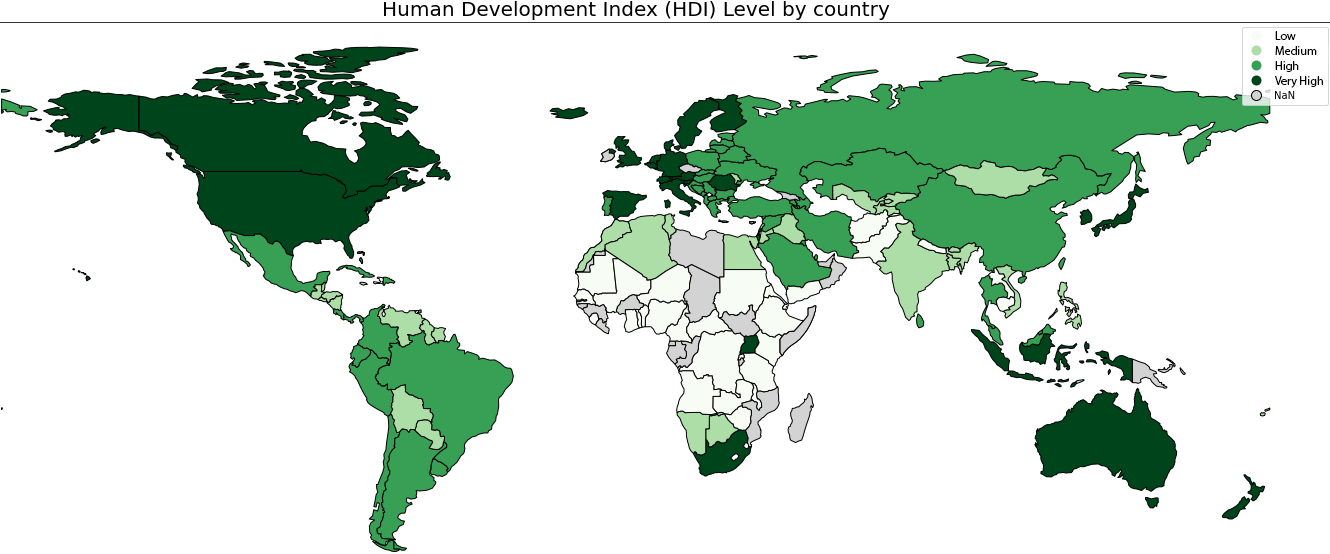}
                \label{fig:hdi-map}
            }
        \subfigure[Cuisine Recipe Helpfulness]  {
                \includegraphics[width=0.48\linewidth]{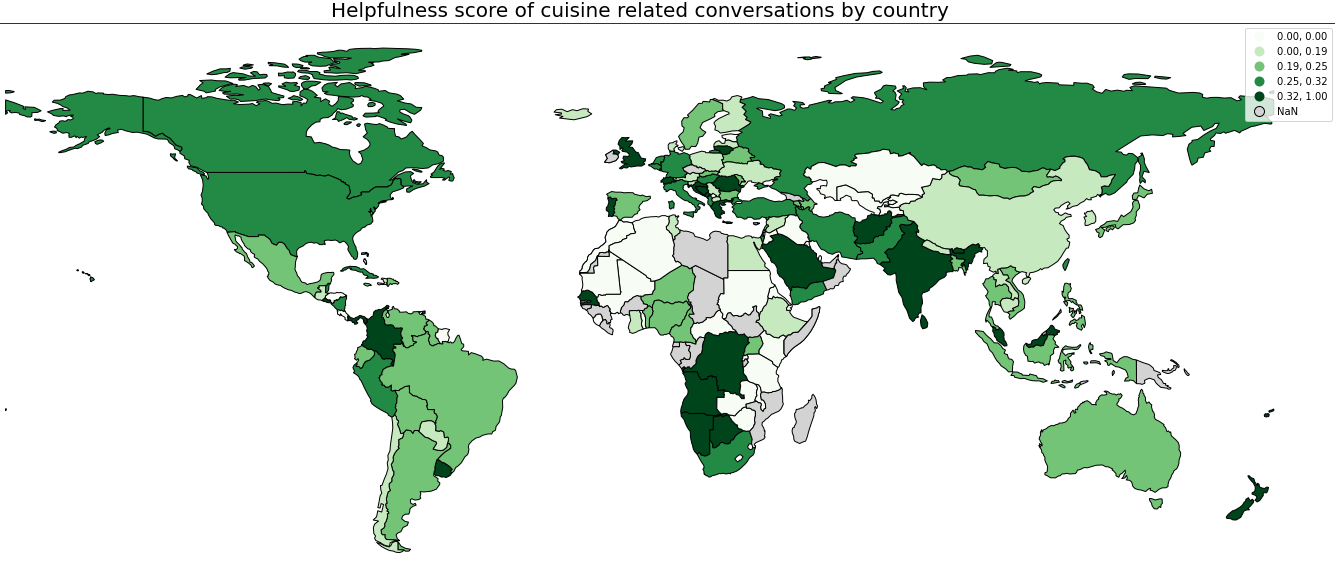}
                \label{fig:holiday}
        }
        \caption{We mark countries with their development level on one map (darker means more developed) and how helpful the GPT-3 model is on another one (darker means more helpful). We see the trend that GPT-3 model tends to be more helpful to more developed countries, uncovering the fairness issue in GPT-3.}
        \label{fig:maps}
\end{figure*} 
After developing the trained helpfulness classifier, we aim to use it to analyze how the model performs under the three scenarios across different countries. As mentioned in Section~\ref{sec:collection}, we use 2,896 (546 cuisine * 4 + 38 holidays * 4 + 140 landmarks * 4) concepts as \emph{analysis examples} in Section~\ref{sec:factual-data}.

\paragraph{GPT-3 and BLENDER.}  We run our trained helpfulness classifier on dialogues generated by GPT-3 and BLENDER on each of those instances. We report the ratio of helpful responses in all generated responses in Table~\ref{tab:fairness}, where we \textbf{boldface} the highest and \underline{underline} the lowest helpfulness across all development levels within each model. We find that GPT-3 and BLENDER tend to be more helpful for questions that contain instances from developed countries than less-developed countries, which indicates a potential \emph{geographical fairness} issue for information-seeking dialogue systems. 

\paragraph{ChatGPT.} We further use ChatGPT~\footnote{We manually query the December 15th (2022) version of ChatGPT research preview and collect ChatGPT's responses.} to generate dialogue responses for all analysis instances and report the model performance as ChatGPT \emph{model} in Table~\ref{tab:fairness}. Two coauthors also manually label the helpfulness of all responses (IAA: 0.74) while collecting the ChatGPT responses, which we report as ChatGPT \emph{Human} in Table~\ref{tab:fairness} for the cuisine scenario. Both model prediction and human annotation show the consensus that ChatGPT, like GPT-3 and BLENDER, also tends to be more helpful for questions that contain instances from developed countries than less-developed countries. Table~\ref{tab:chatgpt} shows the same trends of ChatGPT for the holiday and landmark scenarios. Treating the human annotation as the ground truth, our classifier has F1 scores of 0.98, 0.97, and 0.84 for cooking, landmark, and holiday scenarios separately. This showcases the effectiveness and generalizability of our classifier again.

\paragraph{Helpfulness Breakdown.} We further break down and inspect the three dimensions (i.e., coherent/relevant, useful, informative) for these two dialogue systems across three scenarios. Figure~\ref{fig:breakdown} shows the breakdown analysis. We find that the trend that the model tends to be more helpful for more developed countries stays consistent when looking at each dimension. For example, GPT-3 struggles to even generate relevant and coherent responses for factual information from less developed countries. The conclusion stays the same for BLENDER, which further strengthens our finding that such information-seeking dialogue systems have geographical fairness issues as they have better performance for more developed countries.


\begin{table*}[]
\resizebox{\textwidth}{!}{
\begin{tabular}{@{}lllll@{}}
\toprule
\textbf{\begin{tabular}[c]{@{}c@{}}Develop\end{tabular}} &
  \textbf{Helpful} &
  \multicolumn{1}{c}{\textbf{Query}} &
  \multicolumn{1}{c}{\textbf{GPT-3}} &
  \multicolumn{1}{c}{\textbf{ChatGPT}} \\ \midrule
Low &
  0 &
  \begin{tabular}[c]{@{}l@{}}What can you tell me \\ about the \\ \emph{\textbf{Akachi Monument}}?\end{tabular} &
  In sickness, solve. In health, spare not. &
  \begin{tabular}[c]{@{}l@{}}I'm sorry, but I don't have any information about\\  the Akachi Monument. Could you provide more\\ context or details about what you are looking for?\end{tabular} \\ \midrule
Low &
  0 &
  \begin{tabular}[c]{@{}l@{}}What can you tell me\\ about the\\  \emph{\textbf{Wiichen Men's}}\\\emph{\textbf{Meetinghouse?}}\end{tabular} &
  \begin{tabular}[c]{@{}l@{}}They can teach us many things, but we \\ must work to understand their strange ways.\end{tabular} &
  \begin{tabular}[c]{@{}l@{}}I'm sorry, but I don't have any information about\\ the Wiichen Men's Meetinghouse. Could you \\ provide more context or clarify your question?\end{tabular} \\ \midrule
  \midrule
High &
  1 &
  \begin{tabular}[c]{@{}l@{}}What can you tell me\\ about the\\ \emph{\textbf{Chichen Itza?}}\end{tabular} &
  \begin{tabular}[c]{@{}l@{}}I know that the Chichen Itza is an archaeological\\  site and a monument in Mexico. It's was built by\\ the Maya civilization but no one knows how they\\ created such an amazing city. I know that the \\ Chichen Itza is one of the most visited monuments\\  in Mexico and everyone likes the steps.\end{tabular} &
  \begin{tabular}[c]{@{}l@{}}Chichen Itza is an ancient Mayan city located in\\  the Yucatan Peninsula of Mexico. It is a well-\\ known and well-preserved archaeological site\\ that is visited by thousands of people each year.\\ The city was a major center of political, \\ economic, and religious power in the Mayan \\ civilization and was an important hub of trade\\ and commerce. Chichen Itza is known for its \\ many impressive structures and buildings...\end{tabular} \\ \midrule
Very High &
  1 &
  \begin{tabular}[c]{@{}l@{}}What can you tell me\\ about the\\ \emph{\textbf{Junglinster Longwave}}\\ \emph{\textbf{Transmitter}}?\end{tabular} &
  \begin{tabular}[c]{@{}l@{}}The Junglinster Longwave Transmitter is a high\\ powered frequency-modulated transmitter employed\\  by the Luxembourg government since 1956. The \\ Junglinster Longwave Transmitter operates in the \\ frequency range of 234.7 to 235.3 kHz and has a \\ maximum output power of 405 kW.\end{tabular} &
  \begin{tabular}[c]{@{}l@{}}The Junglinster Longwave Transmitter is a \\ longwave radio transmission facility located\\ near Junglinster, a town in Luxembourg. It is\\ used for long-distance radio communication\\ and broadcasts on the frequency of 153 kHz. \\ The transmitter is operated by the Luxembourg\\ state broadcasting organization, Radio Luxembourg...\end{tabular} \\ \bottomrule
\end{tabular}
}
\caption{Qualitative examples in the landmark scenario used in the fairness analysis. For each query instance, we list the developed level it belongs to (e.g. low-developed, high-developed and very high-developed countries), the helpfulness prediction from the classifier (where 1 for helpful, 0 for not helpful), prompts that we use, and responses generated from GPT-3 and ChatGPT. We admit the potential factual issues and discuss in the Limitations.}
\end{table*}

\section{Related Work}

\paragraph{Dialogue Systems.} \citet{Ni2021RecentAI} comprehensively survey and summarize recent progress in dialogue systems.
For task-oriented dialogue systems (e.g. \citet{NEURIPS2020_e9462095, Yang2021UBARTF}), they focus on accurately handling user's messages for specific tasks such as movie recommendations, 
and successfully finishing the task in limited turns. This type of system usually comes together with databases \cite{yu2021SCoRE} which contain multi-domain knowledge.
Open domain dialogue systems are aiming for coherent and natural conversations \cite{Khatri2018AdvancingTS} that are knowledge equipped \cite{zhao-etal-2020-knowledge-grounded, roller-etal-2021-recipes} or provide empathy \cite{roller-etal-2021-recipes, li-etal-2022-kemp, MA202050} and emotional support \cite{liu-etal-2021-towards}. Sometimes maintaining a consistent persona is also being considered \cite{roller-etal-2021-recipes}. 
While an information-seeking dialogue system \cite{CIS} is the one in between, it aims to provide the help users need.

\paragraph{Dialogue Evaluation.} To evaluate multi-turn task-oriented dialogue systems, the goal success rate \cite{Lu2020EfficientEO, takanobu-etal-2020-goal} is an important metric. 
For single-turn open-domain dialogue, i.e., the agent generates the response based on the given utterance, automatic evaluation is a widely discussed and valuable topic.
~\citet{finch-choi-2020-towards, yeh-etal-2021-comprehensive, Ni2021RecentAI} survey the work in recent years and further analyze and summarize various dialogue evaluation metrics.
Recently, there are more discussions on evaluating dialogue assistants trained with reinforcement learning from human feedback \cite{constitutional, RLHF-train}. Though \citet{HHH-eval} puts forward to evaluate if a language agent is helpful, honest, and harmless, it focuses more on the dialogue alignment perspective.
To our best knowledge, there is no metric to evaluate the helpfulness of single-turn dialogue systems specificly.

\paragraph{Fairness in Dialogue Systems.} Understanding and mitigating societal biases in NLP tasks has been frequently discussed in many recent works. \citet{Dev2021WhatDB} survey existing bias measures and \citet{10.1145/3457607} investigate and categorize fairness and bias in machine learning.
For fairness in language generation, \citet{sheng-etal-2019-woman} systematically evaluate societal biases of text generated from prompts and \citet{Sheng2021SocietalBI} survey and analyze the challenges and progress among dimensions including gender, race, religion, etc. 
For dialogue systems, \citet{Ruane2019ConversationalAS} bring up the social and ethical considerations in conversational agents. \citet{dinan-etal-2020-queens, liu-etal-2020-gender, liu-etal-2020-mitigating} discuss gender bias in dialogue generation and \citet{sheng-etal-2021-nice} investigates the ad hominems in dialogue responses regarding the race perspective. 
As for geographic bias, \citet{jurgens-etal-2017-incorporating, 10.1145/3457607, Suresh2021AFF, yin-etal-2021-broaden} point out the importance of geographic diversity from the data perspective. ~\citet{ghosh-etal-2021-detecting} focus on toxicity detection and center analysis around 7 specific countries, whereas our work looks at dialogue generation models and broadly covers all countries in the world.

\section{Conclusion and Future Works}
Information-seeking dialogue systems have been more and more important and integrated into humans' daily life. 
However, only a few of the previous works has studied how humans perceive the helpfulness of such goal-oriented dialogue systems, let alone the fairness aspect. 
Built on previous works, we propose to evaluate the helpfulness of dialogue systems from three dimensions (relevance \& coherence, usefulness, and informativeness) and collect a large corpus with fine-grained annotations. We use the collected data
to train classifiers that can automatically determine the helpfulness of 
dialogue responses in the single-turn setting.

With the trained classifier, we customize questions and analyze the helpfulness of GPT-3, BLENDER, and ChatGPT in the context of information seeking. Although all three models are known to have Wikipedia knowledge in their training data, they tend to be more helpful for questions asking about instances from more developed countries.  
Such fairness issues could discourage marginalized groups from using these dialogue agents, further reducing
user input to improve dialogue systems. Therefore, we call for imperative attention from the community to carefully examine and address this geographical bias in information-seeking dialogue systems. 


\section*{Limitations}

Our analysis pipeline, including dialogue helpfulness evaluation and fairness analysis, can be generalized to other task-oriented dialogue systems and downstream scenarios. However, one of the limitations of our work is that we have not covered the debiasing method. One promising direction to go is by injecting constraints in the decoding process of generation. However, figuring out how to combine the injection and utilizing knowledge that can help underrepresented groups is challenging. We encourage researchers to build on our work and propose debiasing methods to address the fairness issue in task-oriented dialogue systems. Besides, we do not check the factuality in responses and acknowledge it as one of our limitations. Being closely related to factuality, LLMs may generated hallucinated content. It would also interesting to look into if LLMs' hallucinated content may bias towards certain groups. We leave all these interesting research directions to future works. 



\section*{Ethical Consideration}

Our annotated data come from two sources: Taskmaster-2, an open-source dataset from Google, and GPT-3-, BLENDER- or ChatGPT-generated conversations regarding collected cuisine, holiday, or landmark instances from Wikipedia. 
There is no explicit detail for both sources that leaks information about a user's name, health, negative financial status, racial or ethnic origin, religious or philosophical affiliation, or beliefs.
We also collect crowd-sourced annotations using Amazon Mechanical Turk, where we ask whether a response is helpful 
without collecting information about the annotators.
The annotation information (pay per amount of work, guidelines) is in the appendix, and we ensure the pay per task is above the annotator's local minimum wage.
In addition, we used pre-trained language models (LMs) to generate responses regarding the constructed questions.
Trained on massive online texts, it is well-known that such pretrained LMs could capture the bias reflecting the training data. Therefore, our annotated data for GPT-3 generated responses could also contain offensive content. Interested parties should be careful and inspect them before usage. 

\bibliography{anthology,custom}

\begin{thebibliography}{50}
\expandafter\ifx\csname natexlab\endcsname\relax\def\natexlab#1{#1}\fi

\bibitem[{Askell et~al.(2021)Askell, Bai, Chen, Drain, Ganguli, Henighan, Jones, Joseph, Mann, DasSarma, Elhage, Hatfield-Dodds, Hernandez, Kernion, Ndousse, Olsson, Amodei, Brown, Clark, McCandlish, Olah, and Kaplan}]{HHH-eval}
Amanda Askell, Yuntao Bai, Anna Chen, Dawn Drain, Deep Ganguli, Tom Henighan, Andy Jones, Nicholas Joseph, Ben Mann, Nova DasSarma, Nelson Elhage, Zac Hatfield-Dodds, Danny Hernandez, Jackson Kernion, Kamal Ndousse, Catherine Olsson, Dario Amodei, Tom Brown, Jack Clark, Sam McCandlish, Chris Olah, and Jared Kaplan. 2021.
\newblock \href {https://doi.org/10.48550/ARXIV.2112.00861} {A general language assistant as a laboratory for alignment}.

\bibitem[{Bai et~al.(2022{\natexlab{a}})Bai, Jones, Ndousse, Askell, Chen, DasSarma, Drain, Fort, Ganguli, Henighan, Joseph, Kadavath, Kernion, Conerly, El-Showk, Elhage, Hatfield-Dodds, Hernandez, Hume, Johnston, Kravec, Lovitt, Nanda, Olsson, Amodei, Brown, Clark, McCandlish, Olah, Mann, and Kaplan}]{RLHF-train}
Yuntao Bai, Andy Jones, Kamal Ndousse, Amanda Askell, Anna Chen, Nova DasSarma, Dawn Drain, Stanislav Fort, Deep Ganguli, Tom Henighan, Nicholas Joseph, Saurav Kadavath, Jackson Kernion, Tom Conerly, Sheer El-Showk, Nelson Elhage, Zac Hatfield-Dodds, Danny Hernandez, Tristan Hume, Scott Johnston, Shauna Kravec, Liane Lovitt, Neel Nanda, Catherine Olsson, Dario Amodei, Tom Brown, Jack Clark, Sam McCandlish, Chris Olah, Ben Mann, and Jared Kaplan. 2022{\natexlab{a}}.
\newblock \href {https://doi.org/10.48550/ARXIV.2204.05862} {Training a helpful and harmless assistant with reinforcement learning from human feedback}.

\bibitem[{Bai et~al.(2022{\natexlab{b}})Bai, Kadavath, Kundu, Askell, Kernion, Jones, Chen, Goldie, Mirhoseini, McKinnon, Chen, Olsson, Olah, Hernandez, Drain, Ganguli, Li, Tran-Johnson, Perez, Kerr, Mueller, Ladish, Landau, Ndousse, Lukosuite, Lovitt, Sellitto, Elhage, Schiefer, Mercado, DasSarma, Lasenby, Larson, Ringer, Johnston, Kravec, Showk, Fort, Lanham, Telleen-Lawton, Conerly, Henighan, Hume, Bowman, Hatfield-Dodds, Mann, Amodei, Joseph, McCandlish, Brown, and Kaplan}]{constitutional}
Yuntao Bai, Saurav Kadavath, Sandipan Kundu, Amanda Askell, Jackson Kernion, Andy Jones, Anna Chen, Anna Goldie, Azalia Mirhoseini, Cameron McKinnon, Carol Chen, Catherine Olsson, Christopher Olah, Danny Hernandez, Dawn Drain, Deep Ganguli, Dustin Li, Eli Tran-Johnson, Ethan Perez, Jamie Kerr, Jared Mueller, Jeffrey Ladish, Joshua Landau, Kamal Ndousse, Kamile Lukosuite, Liane Lovitt, Michael Sellitto, Nelson Elhage, Nicholas Schiefer, Noemi Mercado, Nova DasSarma, Robert Lasenby, Robin Larson, Sam Ringer, Scott Johnston, Shauna Kravec, Sheer~El Showk, Stanislav Fort, Tamera Lanham, Timothy Telleen-Lawton, Tom Conerly, Tom Henighan, Tristan Hume, Samuel~R. Bowman, Zac Hatfield-Dodds, Ben Mann, Dario Amodei, Nicholas Joseph, Sam McCandlish, Tom Brown, and Jared Kaplan. 2022{\natexlab{b}}.
\newblock \href {https://doi.org/10.48550/ARXIV.2212.08073} {Constitutional ai: Harmlessness from ai feedback}.

\bibitem[{Brown et~al.(2020)Brown, Mann, Ryder, Subbiah, Kaplan, Dhariwal, Neelakantan, Shyam, Sastry, Askell, Agarwal, Herbert{-}Voss, Krueger, Henighan, Child, Ramesh, Ziegler, Wu, Winter, Hesse, Chen, Sigler, Litwin, Gray, Chess, Clark, Berner, McCandlish, Radford, Sutskever, and Amodei}]{gpt3}
Tom~B. Brown, Benjamin Mann, Nick Ryder, Melanie Subbiah, Jared Kaplan, Prafulla Dhariwal, Arvind Neelakantan, Pranav Shyam, Girish Sastry, Amanda Askell, Sandhini Agarwal, Ariel Herbert{-}Voss, Gretchen Krueger, Tom Henighan, Rewon Child, Aditya Ramesh, Daniel~M. Ziegler, Jeffrey Wu, Clemens Winter, Christopher Hesse, Mark Chen, Eric Sigler, Mateusz Litwin, Scott Gray, Benjamin Chess, Jack Clark, Christopher Berner, Sam McCandlish, Alec Radford, Ilya Sutskever, and Dario Amodei. 2020.
\newblock \href {http://arxiv.org/abs/2005.14165} {Language models are few-shot learners}.
\newblock \emph{CoRR}, abs/2005.14165.

\bibitem[{Byrne et~al.(2021)Byrne, Krishnamoorthi, Ganesh, and Kale}]{byrne-etal-2021-tickettalk}
Bill Byrne, Karthik Krishnamoorthi, Saravanan Ganesh, and Mihir Kale. 2021.
\newblock \href {https://doi.org/10.18653/v1/2021.acl-long.55} {{T}icket{T}alk: Toward human-level performance with end-to-end, transaction-based dialog systems}.
\newblock In \emph{Proceedings of the 59th Annual Meeting of the Association for Computational Linguistics and the 11th International Joint Conference on Natural Language Processing (Volume 1: Long Papers)}, pages 671--680, Online. Association for Computational Linguistics.

\bibitem[{Byrne et~al.(2019)Byrne, Krishnamoorthi, Sankar, Neelakantan, Goodrich, Duckworth, Yavuz, Dubey, Kim, and Cedilnik}]{byrne-etal-2019-taskmaster}
Bill Byrne, Karthik Krishnamoorthi, Chinnadhurai Sankar, Arvind Neelakantan, Ben Goodrich, Daniel Duckworth, Semih Yavuz, Amit Dubey, Kyu-Young Kim, and Andy Cedilnik. 2019.
\newblock \href {https://doi.org/10.18653/v1/D19-1459} {Taskmaster-1: Toward a realistic and diverse dialog dataset}.
\newblock In \emph{Proceedings of the 2019 Conference on Empirical Methods in Natural Language Processing and the 9th International Joint Conference on Natural Language Processing (EMNLP-IJCNLP)}, pages 4516--4525, Hong Kong, China. Association for Computational Linguistics.

\bibitem[{Delobelle et~al.(2020)Delobelle, Winters, and Berendt}]{delobelle-etal-2020-robbert}
Pieter Delobelle, Thomas Winters, and Bettina Berendt. 2020.
\newblock \href {https://doi.org/10.18653/v1/2020.findings-emnlp.292} {{R}ob{BERT}: a {D}utch {R}o{BERT}a-based {L}anguage {M}odel}.
\newblock In \emph{Findings of the Association for Computational Linguistics: EMNLP 2020}, pages 3255--3265, Online. Association for Computational Linguistics.

\bibitem[{Dev et~al.(2021)Dev, Sheng, Zhao, Sun, Hou, Sanseverino, Kim, Peng, and Chang}]{Dev2021WhatDB}
Sunipa Dev, Emily Sheng, Jieyu Zhao, Jiao Sun, Yu~Hou, Mattie Sanseverino, Jiin Kim, Nanyun Peng, and Kai-Wei Chang. 2021.
\newblock What do bias measures measure?
\newblock \emph{ArXiv}, abs/2108.03362.

\bibitem[{Devlin et~al.(2019)Devlin, Chang, Lee, and Toutanova}]{devlin-etal-2019-bert}
Jacob Devlin, Ming-Wei Chang, Kenton Lee, and Kristina Toutanova. 2019.
\newblock \href {https://doi.org/10.18653/v1/N19-1423} {{BERT}: Pre-training of deep bidirectional transformers for language understanding}.
\newblock In \emph{Proceedings of the 2019 Conference of the North {A}merican Chapter of the Association for Computational Linguistics: Human Language Technologies, Volume 1 (Long and Short Papers)}, pages 4171--4186, Minneapolis, Minnesota. Association for Computational Linguistics.

\bibitem[{Dinan et~al.(2020)Dinan, Fan, Williams, Urbanek, Kiela, and Weston}]{dinan-etal-2020-queens}
Emily Dinan, Angela Fan, Adina Williams, Jack Urbanek, Douwe Kiela, and Jason Weston. 2020.
\newblock \href {https://doi.org/10.18653/v1/2020.emnlp-main.656} {Queens are powerful too: Mitigating gender bias in dialogue generation}.
\newblock In \emph{Proceedings of the 2020 Conference on Empirical Methods in Natural Language Processing (EMNLP)}, pages 8173--8188, Online. Association for Computational Linguistics.

\bibitem[{Finch and Choi(2020)}]{finch-choi-2020-towards}
Sarah~E. Finch and Jinho~D. Choi. 2020.
\newblock \href {https://aclanthology.org/2020.sigdial-1.29} {Towards unified dialogue system evaluation: A comprehensive analysis of current evaluation protocols}.
\newblock In \emph{Proceedings of the 21th Annual Meeting of the Special Interest Group on Discourse and Dialogue}, pages 236--245, 1st virtual meeting. Association for Computational Linguistics.

\bibitem[{Fleiss and Cohen(1973)}]{Fleiss1973TheEO}
Joseph~L. Fleiss and Jacob Cohen. 1973.
\newblock The equivalence of weighted kappa and the intraclass correlation coefficient as measures of reliability.
\newblock \emph{Educational and Psychological Measurement}, 33:613 -- 619.

\bibitem[{Ghosh et~al.(2021)Ghosh, Baker, Jurgens, and Prabhakaran}]{ghosh-etal-2021-detecting}
Sayan Ghosh, Dylan Baker, David Jurgens, and Vinodkumar Prabhakaran. 2021.
\newblock \href {https://aclanthology.org/2021.wnut-1.35} {Detecting cross-geographic biases in toxicity modeling on social media}.
\newblock In \emph{Proceedings of the Seventh Workshop on Noisy User-generated Text (W-NUT 2021)}, pages 313--328, Online. Association for Computational Linguistics.

\bibitem[{Glaese et~al.(2022)Glaese, McAleese, Trębacz, Aslanides, Firoiu, Ewalds, Rauh, Weidinger, Chadwick, Thacker, Campbell-Gillingham, Uesato, Huang, Comanescu, Yang, See, Dathathri, Greig, Chen, Fritz, Elias, Green, Mokrá, Fernando, Wu, Foley, Young, Gabriel, Isaac, Mellor, Hassabis, Kavukcuoglu, Hendricks, and Irving}]{sparrow}
Amelia Glaese, Nat McAleese, Maja Trębacz, John Aslanides, Vlad Firoiu, Timo Ewalds, Maribeth Rauh, Laura Weidinger, Martin Chadwick, Phoebe Thacker, Lucy Campbell-Gillingham, Jonathan Uesato, Po-Sen Huang, Ramona Comanescu, Fan Yang, Abigail See, Sumanth Dathathri, Rory Greig, Charlie Chen, Doug Fritz, Jaume~Sanchez Elias, Richard Green, Soňa Mokrá, Nicholas Fernando, Boxi Wu, Rachel Foley, Susannah Young, Iason Gabriel, William Isaac, John Mellor, Demis Hassabis, Koray Kavukcuoglu, Lisa~Anne Hendricks, and Geoffrey Irving. 2022.
\newblock \href {https://doi.org/10.48550/ARXIV.2209.14375} {Improving alignment of dialogue agents via targeted human judgements}.

\bibitem[{He et~al.(2021)He, Liu, Gao, and Chen}]{deberta}
Pengcheng He, Xiaodong Liu, Jianfeng Gao, and Weizhu Chen. 2021.
\newblock Deberta: Decoding-enhanced bert with disentangled attention.
\newblock \emph{ArXiv}, abs/2006.03654.

\bibitem[{Hosseini-Asl et~al.(2020)Hosseini-Asl, McCann, Wu, Yavuz, and Socher}]{NEURIPS2020_e9462095}
Ehsan Hosseini-Asl, Bryan McCann, Chien-Sheng Wu, Semih Yavuz, and Richard Socher. 2020.
\newblock \href {https://proceedings.neurips.cc/paper/2020/file/e946209592563be0f01c844ab2170f0c-Paper.pdf} {A simple language model for task-oriented dialogue}.
\newblock In \emph{Advances in Neural Information Processing Systems}, volume~33, pages 20179--20191. Curran Associates, Inc.

\bibitem[{Jeni et~al.(2013)Jeni, Cohn, and De~La~Torre}]{jeni2013facing}
L{\'a}szl{\'o}~A Jeni, Jeffrey~F Cohn, and Fernando De~La~Torre. 2013.
\newblock Facing imbalanced data--recommendations for the use of performance metrics.
\newblock In \emph{2013 Humaine association conference on affective computing and intelligent interaction}, pages 245--251. IEEE.

\bibitem[{Jurgens et~al.(2017)Jurgens, Tsvetkov, and Jurafsky}]{jurgens-etal-2017-incorporating}
David Jurgens, Yulia Tsvetkov, and Dan Jurafsky. 2017.
\newblock \href {https://doi.org/10.18653/v1/P17-2009} {Incorporating dialectal variability for socially equitable language identification}.
\newblock In \emph{Proceedings of the 55th Annual Meeting of the Association for Computational Linguistics (Volume 2: Short Papers)}, pages 51--57, Vancouver, Canada. Association for Computational Linguistics.

\bibitem[{Khatri et~al.(2018)Khatri, Hedayatnia, Venkatesh, Nunn, Pan, Liu, Song, Gottardi, Kwatra, Pancholi, Cheng, Chen, Stubel, Gopalakrishnan, Bland, Gabriel, Mandal, Hakkani-T{\"u}r, Hwang, Michel, King, and Prasad}]{Khatri2018AdvancingTS}
Chandra Khatri, Behnam Hedayatnia, Anu Venkatesh, Jeff Nunn, Yi~Pan, Qing Liu, Han Song, Anna Gottardi, Sanjeev Kwatra, Sanju Pancholi, Ming Cheng, Qinglang Chen, Lauren Stubel, Karthik Gopalakrishnan, Kate Bland, Raefer Gabriel, Arindam Mandal, Dilek~Z. Hakkani-T{\"u}r, Gene Hwang, Nate Michel, Eric King, and Rohit Prasad. 2018.
\newblock Advancing the state of the art in open domain dialog systems through the alexa prize.
\newblock \emph{ArXiv}, abs/1812.10757.

\bibitem[{Lan et~al.(2020)Lan, Chen, Goodman, Gimpel, Sharma, and Soricut}]{albert}
Zhenzhong Lan, Mingda Chen, Sebastian Goodman, Kevin Gimpel, Piyush Sharma, and Radu Soricut. 2020.
\newblock Albert: A lite bert for self-supervised learning of language representations.
\newblock \emph{ArXiv}, abs/1909.11942.

\bibitem[{Li et~al.(2022)Li, Li, Ren, Ren, and Chen}]{li-etal-2022-kemp}
Qintong Li, Piji Li, Zhaochun Ren, Pengjie Ren, and Zhumin Chen. 2022.
\newblock Knowledge bridging for empathetic dialogue generation.

\bibitem[{Liu et~al.(2020{\natexlab{a}})Liu, Dacon, Fan, Liu, Liu, and Tang}]{liu-etal-2020-gender}
Haochen Liu, Jamell Dacon, Wenqi Fan, Hui Liu, Zitao Liu, and Jiliang Tang. 2020{\natexlab{a}}.
\newblock \href {https://doi.org/10.18653/v1/2020.coling-main.390} {Does gender matter? towards fairness in dialogue systems}.
\newblock In \emph{Proceedings of the 28th International Conference on Computational Linguistics}, pages 4403--4416, Barcelona, Spain (Online). International Committee on Computational Linguistics.

\bibitem[{Liu et~al.(2020{\natexlab{b}})Liu, Wang, Wang, Liu, Liu, and Tang}]{liu-etal-2020-mitigating}
Haochen Liu, Wentao Wang, Yiqi Wang, Hui Liu, Zitao Liu, and Jiliang Tang. 2020{\natexlab{b}}.
\newblock \href {https://doi.org/10.18653/v1/2020.emnlp-main.64} {Mitigating gender bias for neural dialogue generation with adversarial learning}.
\newblock In \emph{Proceedings of the 2020 Conference on Empirical Methods in Natural Language Processing (EMNLP)}, pages 893--903, Online. Association for Computational Linguistics.

\bibitem[{Liu et~al.(2021)Liu, Zheng, Demasi, Sabour, Li, Yu, Jiang, and Huang}]{liu-etal-2021-towards}
Siyang Liu, Chujie Zheng, Orianna Demasi, Sahand Sabour, Yu~Li, Zhou Yu, Yong Jiang, and Minlie Huang. 2021.
\newblock \href {https://doi.org/10.18653/v1/2021.acl-long.269} {Towards emotional support dialog systems}.
\newblock In \emph{Proceedings of the 59th Annual Meeting of the Association for Computational Linguistics and the 11th International Joint Conference on Natural Language Processing (Volume 1: Long Papers)}, pages 3469--3483, Online. Association for Computational Linguistics.

\bibitem[{Liu et~al.(2023)Liu, Iter, Xu, Wang, Xu, and Zhu}]{liu2023g-eval}
Yang Liu, Dan Iter, Yichong Xu, Shuohang Wang, Ruochen Xu, and Chenguang Zhu. 2023.
\newblock \href {https://www.microsoft.com/en-us/research/publication/gpteval-nlg-evaluation-using-gpt-4-with-better-human-alignment/} {G-eval: Nlg evaluation using gpt-4 with better human alignment}.
\newblock \emph{arXiv 2303.16634}.

\bibitem[{Lu et~al.(2020)Lu, Xu, and Li}]{Lu2020EfficientEO}
Weiyi Lu, Yi~Xu, and Li~Erran Li. 2020.
\newblock Efficient evaluation of task oriented dialogue systems.

\bibitem[{Ma et~al.(2020)Ma, Nguyen, Xing, and Cambria}]{MA202050}
Yukun Ma, Khanh~Linh Nguyen, Frank~Z. Xing, and Erik Cambria. 2020.
\newblock \href {https://doi.org/https://doi.org/10.1016/j.inffus.2020.06.011} {A survey on empathetic dialogue systems}.
\newblock \emph{Information Fusion}, 64:50--70.

\bibitem[{McHugh(2012)}]{mchugh2012interrater}
Mary~Lou McHugh. 2012.
\newblock Interrater reliability: The kappa statistic.
\newblock \emph{Biochem Med (Zagreb)}, 22(3):276--82.

\bibitem[{Mehrabi et~al.(2021)Mehrabi, Morstatter, Saxena, Lerman, and Galstyan}]{10.1145/3457607}
Ninareh Mehrabi, Fred Morstatter, Nripsuta Saxena, Kristina Lerman, and Aram Galstyan. 2021.
\newblock \href {https://doi.org/10.1145/3457607} {A survey on bias and fairness in machine learning}.
\newblock \emph{ACM Comput. Surv.}, 54(6).

\bibitem[{Ni et~al.(2021)Ni, Young, Pandelea, Xue, Adiga, and Cambria}]{Ni2021RecentAI}
Jinjie Ni, Tom Young, Vlad Pandelea, Fuzhao Xue, V.~Ananth~Krishna Adiga, and E.~Cambria. 2021.
\newblock Recent advances in deep learning based dialogue systems: A systematic survey.
\newblock \emph{ArXiv}, abs/2105.04387.

\bibitem[{Ning et~al.(2020)Ning, Wu, Dasigi, Dua, Gardner, Logan~IV, Marasovi{\'c}, and Nie}]{ning-etal-2020-easy}
Qiang Ning, Hao Wu, Pradeep Dasigi, Dheeru Dua, Matt Gardner, Robert~L. Logan~IV, Ana Marasovi{\'c}, and Zhen Nie. 2020.
\newblock \href {https://doi.org/10.18653/v1/2020.emnlp-demos.17} {Easy, reproducible and quality-controlled data collection with {CROWDAQ}}.
\newblock In \emph{Proceedings of the 2020 Conference on Empirical Methods in Natural Language Processing: System Demonstrations}, pages 127--134, Online. Association for Computational Linguistics.

\bibitem[{Ram et~al.(2018)Ram, Prasad, Khatri, Venkatesh, Gabriel, Liu, Nunn, Hedayatnia, Cheng, Nagar, King, Bland, Wartick, Pan, Song, Jayadevan, Hwang, and Pettigrue}]{Ram2018ConversationalAT}
Ashwin Ram, Rohit Prasad, Chandra Khatri, Anu Venkatesh, Raefer Gabriel, Qing Liu, Jeff Nunn, Behnam Hedayatnia, Ming Cheng, Ashish Nagar, Eric King, Kate Bland, Amanda Wartick, Yi~Pan, Han Song, Sk~Jayadevan, Gene Hwang, and Art Pettigrue. 2018.
\newblock Conversational ai: The science behind the alexa prize.
\newblock \emph{ArXiv}, abs/1801.03604.

\bibitem[{Roller et~al.(2021)Roller, Dinan, Goyal, Ju, Williamson, Liu, Xu, Ott, Smith, Boureau, and Weston}]{roller-etal-2021-recipes}
Stephen Roller, Emily Dinan, Naman Goyal, Da~Ju, Mary Williamson, Yinhan Liu, Jing Xu, Myle Ott, Eric~Michael Smith, Y-Lan Boureau, and Jason Weston. 2021.
\newblock \href {https://doi.org/10.18653/v1/2021.eacl-main.24} {Recipes for building an open-domain chatbot}.
\newblock In \emph{Proceedings of the 16th Conference of the European Chapter of the Association for Computational Linguistics: Main Volume}, pages 300--325, Online. Association for Computational Linguistics.

\bibitem[{Ruane et~al.(2019)Ruane, Birhane, and Ventresque}]{Ruane2019ConversationalAS}
Elayne Ruane, Abeba Birhane, and Anthony Ventresque. 2019.
\newblock Conversational ai: Social and ethical considerations.
\newblock In \emph{AICS}.

\bibitem[{Sheng et~al.(2021{\natexlab{a}})Sheng, Chang, Natarajan, and Peng}]{Sheng2021SocietalBI}
Emily Sheng, Kai-Wei Chang, P.~Natarajan, and Nanyun Peng. 2021{\natexlab{a}}.
\newblock Societal biases in language generation: Progress and challenges.
\newblock In \emph{ACL/IJCNLP}.

\bibitem[{Sheng et~al.(2021{\natexlab{b}})Sheng, Chang, Natarajan, and Peng}]{sheng-etal-2021-nice}
Emily Sheng, Kai-Wei Chang, Prem Natarajan, and Nanyun Peng. 2021{\natexlab{b}}.
\newblock \href {https://doi.org/10.18653/v1/2021.naacl-main.60} {{``}nice try, kiddo{''}: Investigating ad hominems in dialogue responses}.
\newblock In \emph{Proceedings of the 2021 Conference of the North American Chapter of the Association for Computational Linguistics: Human Language Technologies}, pages 750--767, Online. Association for Computational Linguistics.

\bibitem[{Sheng et~al.(2019)Sheng, Chang, Natarajan, and Peng}]{sheng-etal-2019-woman}
Emily Sheng, Kai-Wei Chang, Premkumar Natarajan, and Nanyun Peng. 2019.
\newblock \href {https://doi.org/10.18653/v1/D19-1339} {The woman worked as a babysitter: On biases in language generation}.
\newblock In \emph{Proceedings of the 2019 Conference on Empirical Methods in Natural Language Processing and the 9th International Joint Conference on Natural Language Processing (EMNLP-IJCNLP)}, pages 3407--3412, Hong Kong, China. Association for Computational Linguistics.

\bibitem[{Suresh and Guttag(2021)}]{Suresh2021AFF}
Harini Suresh and John~V. Guttag. 2021.
\newblock A framework for understanding sources of harm throughout the machine learning life cycle.
\newblock \emph{Equity and Access in Algorithms, Mechanisms, and Optimization}.

\bibitem[{Takanobu et~al.(2020)Takanobu, Zhu, Li, Peng, Gao, and Huang}]{takanobu-etal-2020-goal}
Ryuichi Takanobu, Qi~Zhu, Jinchao Li, Baolin Peng, Jianfeng Gao, and Minlie Huang. 2020.
\newblock \href {https://aclanthology.org/2020.sigdial-1.37} {Is your goal-oriented dialog model performing really well? empirical analysis of system-wise evaluation}.
\newblock In \emph{Proceedings of the 21th Annual Meeting of the Special Interest Group on Discourse and Dialogue}, pages 297--310, 1st virtual meeting. Association for Computational Linguistics.

\bibitem[{Thoppilan et~al.(2022)Thoppilan, De~Freitas, Hall, Shazeer, Kulshreshtha, Cheng, Jin, Bos, Baker, Du, Li, Lee, Zheng, Ghafouri, Menegali, Huang, Krikun, Lepikhin, Qin, Chen, Xu, Chen, Roberts, Bosma, Zhao, Zhou, Chang, Krivokon, Rusch, Pickett, Srinivasan, Man, Meier-Hellstern, Morris, Doshi, Santos, Duke, Soraker, Zevenbergen, Prabhakaran, Diaz, Hutchinson, Olson, Molina, Hoffman-John, Lee, Aroyo, Rajakumar, Butryna, Lamm, Kuzmina, Fenton, Cohen, Bernstein, Kurzweil, Aguera-Arcas, Cui, Croak, Chi, and Le}]{lamda}
Romal Thoppilan, Daniel De~Freitas, Jamie Hall, Noam Shazeer, Apoorv Kulshreshtha, Heng-Tze Cheng, Alicia Jin, Taylor Bos, Leslie Baker, Yu~Du, YaGuang Li, Hongrae Lee, Huaixiu~Steven Zheng, Amin Ghafouri, Marcelo Menegali, Yanping Huang, Maxim Krikun, Dmitry Lepikhin, James Qin, Dehao Chen, Yuanzhong Xu, Zhifeng Chen, Adam Roberts, Maarten Bosma, Vincent Zhao, Yanqi Zhou, Chung-Ching Chang, Igor Krivokon, Will Rusch, Marc Pickett, Pranesh Srinivasan, Laichee Man, Kathleen Meier-Hellstern, Meredith~Ringel Morris, Tulsee Doshi, Renelito~Delos Santos, Toju Duke, Johnny Soraker, Ben Zevenbergen, Vinodkumar Prabhakaran, Mark Diaz, Ben Hutchinson, Kristen Olson, Alejandra Molina, Erin Hoffman-John, Josh Lee, Lora Aroyo, Ravi Rajakumar, Alena Butryna, Matthew Lamm, Viktoriya Kuzmina, Joe Fenton, Aaron Cohen, Rachel Bernstein, Ray Kurzweil, Blaise Aguera-Arcas, Claire Cui, Marian Croak, Ed~Chi, and Quoc Le. 2022.
\newblock \href {https://doi.org/10.48550/ARXIV.2201.08239} {Lamda: Language models for dialog applications}.

\bibitem[{Voorhees(2008)}]{Voorhees2008EvaluatingQA}
Ellen~M. Voorhees. 2008.
\newblock Evaluating question answering system performance.

\bibitem[{Wang et~al.(2023)Wang, Li, Chen, Zhu, Lin, Cao, Liu, Liu, and Sui}]{wang2023large}
Peiyi Wang, Lei Li, Liang Chen, Dawei Zhu, Binghuai Lin, Yunbo Cao, Qi~Liu, Tianyu Liu, and Zhifang Sui. 2023.
\newblock \href {http://arxiv.org/abs/2305.17926} {Large language models are not fair evaluators}.

\bibitem[{Wen et~al.(2017)Wen, Vandyke, Mrk{\v{s}}i{\'c}, Ga{\v{s}}i{\'c}, Rojas-Barahona, Su, Ultes, and Young}]{wen-etal-2017-network}
Tsung-Hsien Wen, David Vandyke, Nikola Mrk{\v{s}}i{\'c}, Milica Ga{\v{s}}i{\'c}, Lina~M. Rojas-Barahona, Pei-Hao Su, Stefan Ultes, and Steve Young. 2017.
\newblock \href {https://aclanthology.org/E17-1042} {A network-based end-to-end trainable task-oriented dialogue system}.
\newblock In \emph{Proceedings of the 15th Conference of the {E}uropean Chapter of the Association for Computational Linguistics: Volume 1, Long Papers}, pages 438--449, Valencia, Spain. Association for Computational Linguistics.

\bibitem[{Wolf et~al.(2020)Wolf, Debut, Sanh, Chaumond, Delangue, Moi, Cistac, Rault, Louf, Funtowicz, Davison, Shleifer, von Platen, Ma, Jernite, Plu, Xu, Le~Scao, Gugger, Drame, Lhoest, and Rush}]{wolf-etal-2020-transformers}
Thomas Wolf, Lysandre Debut, Victor Sanh, Julien Chaumond, Clement Delangue, Anthony Moi, Pierric Cistac, Tim Rault, Remi Louf, Morgan Funtowicz, Joe Davison, Sam Shleifer, Patrick von Platen, Clara Ma, Yacine Jernite, Julien Plu, Canwen Xu, Teven Le~Scao, Sylvain Gugger, Mariama Drame, Quentin Lhoest, and Alexander Rush. 2020.
\newblock \href {https://doi.org/10.18653/v1/2020.emnlp-demos.6} {Transformers: State-of-the-art natural language processing}.
\newblock In \emph{Proceedings of the 2020 Conference on Empirical Methods in Natural Language Processing: System Demonstrations}, pages 38--45, Online. Association for Computational Linguistics.

\bibitem[{Yang et~al.(2021)Yang, Li, and Quan}]{Yang2021UBARTF}
Yunyi Yang, Yunhao Li, and Xiaojun Quan. 2021.
\newblock Ubar: Towards fully end-to-end task-oriented dialog systems with gpt-2.
\newblock In \emph{AAAI}.

\bibitem[{Yeh et~al.(2021)Yeh, Eskenazi, and Mehri}]{yeh-etal-2021-comprehensive}
Yi-Ting Yeh, Maxine Eskenazi, and Shikib Mehri. 2021.
\newblock \href {https://aclanthology.org/2021.eancs-1.3} {A comprehensive assessment of dialog evaluation metrics}.
\newblock In \emph{The First Workshop on Evaluations and Assessments of Neural Conversation Systems}, pages 15--33, Online. Association for Computational Linguistics.

\bibitem[{Yin et~al.(2021)Yin, Li, Hu, Peng, and Chang}]{yin-etal-2021-broaden}
Da~Yin, Liunian~Harold Li, Ziniu Hu, Nanyun Peng, and Kai-Wei Chang. 2021.
\newblock \href {https://aclanthology.org/2021.emnlp-main.162} {Broaden the vision: Geo-diverse visual commonsense reasoning}.
\newblock In \emph{Proceedings of the 2021 Conference on Empirical Methods in Natural Language Processing}, pages 2115--2129, Online and Punta Cana, Dominican Republic. Association for Computational Linguistics.

\bibitem[{Yu et~al.(2021)Yu, Zhang, Polozov, Meek, and Awadallah}]{yu2021SCoRE}
Tao Yu, Rui Zhang, Oleksandr Polozov, Christopher Meek, and Ahmed~Hassan Awadallah. 2021.
\newblock \href {https://openreview.net/forum?id=oyZxhRI2RiE} {{SCoRE}: Pre-training for context representation in conversational semantic parsing}.
\newblock In \emph{International Conference on Learning Representations}.

\bibitem[{Zamani et~al.(2022)Zamani, Trippas, Dalton, and Radlinski}]{CIS}
Hamed Zamani, Johanne~R. Trippas, Jeff Dalton, and Filip Radlinski. 2022.
\newblock \href {https://doi.org/10.48550/ARXIV.2201.08808} {Conversational information seeking}.

\bibitem[{Zhao et~al.(2020)Zhao, Wu, Xu, Tao, Zhao, and Yan}]{zhao-etal-2020-knowledge-grounded}
Xueliang Zhao, Wei Wu, Can Xu, Chongyang Tao, Dongyan Zhao, and Rui Yan. 2020.
\newblock \href {https://doi.org/10.18653/v1/2020.emnlp-main.272} {Knowledge-grounded dialogue generation with pre-trained language models}.
\newblock In \emph{Proceedings of the 2020 Conference on Empirical Methods in Natural Language Processing (EMNLP)}, pages 3377--3390, Online. Association for Computational Linguistics.

\end{thebibliography}
\bibliographystyle{acl_natbib}

\clearpage
\appendix
\section{Annotation Details}
\label{app:annotation}

\paragraph{Qualification Quiz} To familiarize the workers with the annotation task and get qualified workers, we first conduct a qualification quiz on MTurk using CROWDAQ \cite{ning-etal-2020-easy}.
As for the tutorial of the quiz, we provide 24 dialogue annotation examples among relevant \& coherent, useful and informative dimensions.
Then we ask workers to annotate 10 dialogues, i.e. 30 multi-choice questions to test their understanding. 
To pass the test, they need at least answer 24 questions correctly (i.e., the accuracy is equal or higher than 80\%).
The base payment of the quiz is \$0.1 but people who pass it will earn \$1 bonus.
Finally, there are 44 workers with a \textit{HIT Approval Rate} greater than 98\% and the \textit{Number of HITs Approved} greater than 5000 in our qualified work pool, and they will be invited for each round of the helpfulness annotation task.  

\begin{figure*}
    \centering
    \includegraphics[width=0.9\linewidth]{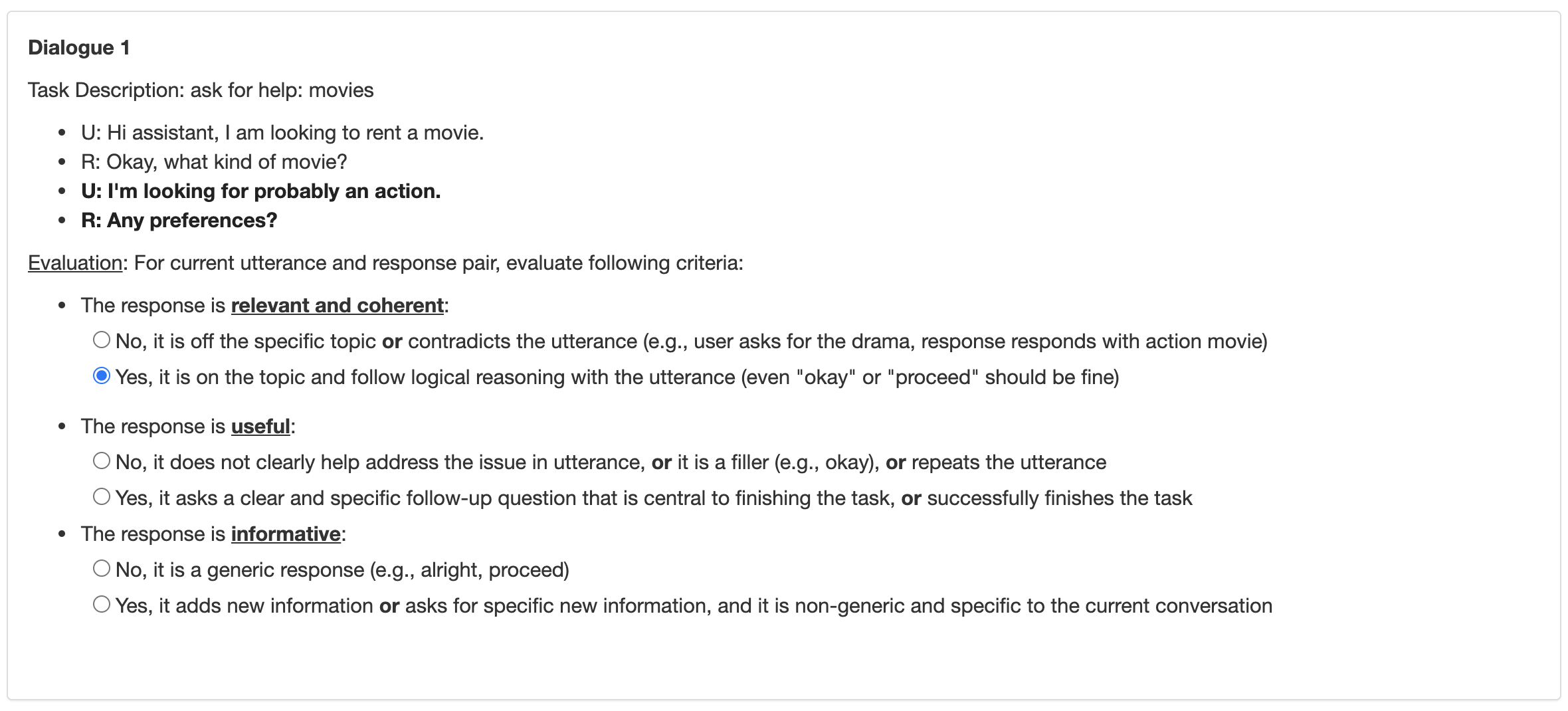}
    \caption{MTurk annotation interface of annotating a dialogue.}
    \label{fig:app-annotation}
\end{figure*}

\paragraph{Annotation Task} Our task involves three parts in total: Define Helpfulness, Annotation Tutorial and Advice on Definition.
To define what helpfulness in single-turn conversations is, we first ask workers their own helpfulness criteria and then share our definition.
Next, we guide them how to do the two-step annotation for each dialogue with illustrations and examples. For each HIT, there are 10 dialogues. Figure~\ref{fig:app-annotation} is the screenshot of how the annotation task looks like.
After they finish the annotation, we ask them again if they have more dimensions to evaluate the helpfulness.
To encourage workers answer our free-form open questions, we provide extra bonus for high-quality answers.
As our task is pretty easy and quick to finish, the payment for each HIT is \$ 0.5. 
We explain how the data would be used at the beginning of the instructions and we this research has been reviewed by the \textbf{IRB board} and granted the status of an \textbf{IRB exempt}.

\paragraph{Helpfulness Criteria} Here are the relevant \& coherent, useful and informative criteria we use to instruct workers. \textbf{U} stands for \textit{Utterance}, \textbf{R} is for \textit{Response} and \textbf{E} is for \textit{Explanation}.
\begin{itemize}[leftmargin=*]
\itemsep-.5em 
\vspace{-0.25em}
    \item Relevant \& Coherent
    \begin{itemize}[leftmargin=*]
        \item \textbf{Criteria}: It measures the collaboration and control over the topic of a conversation. The response is relevant and coherent if it is on-topic with the immediate dialogue history and follows logical reasoning throughout the whole conversation. The response can be considered incoherent when it directly contradicts its previous utterances.
        \item \textbf{Rule of thumb}: The response should be on-topic and follow logical reasoning with the utterance (even "okay" or "proceed" should be fine).
        \item Bad example 1:
        \begin{itemize}
            \item[\textbf{U:}] What should I do when a random woman approaches me?
            \item[\textbf{R:}] Usually, the women select the men based on their beauty, height, and wealth. If you want to find a woman for you, find yourself a pretty woman.
            \item[\textbf{E:}] The user is asking about the actions they need to take, but the answer is irrelevant and incoherent to the topic.
        \end{itemize}
        \item Bad example 2:
        \begin{itemize}
            \item[\textbf{U:}] I want to find a comedy.
            \item[\textbf{R:}] Sure. I have some best thrillers for you.
            \item[\textbf{E:}] The utterance asks for a comedy but the response gives thrillers, which is irrelevant and incoherent to the topic.
        \end{itemize}
    \end{itemize}
    \vspace{0.3em}
    \item Useful
    \begin{itemize}[leftmargin=*]
        \item \textbf{Criteria}: If the utterance is goal-oriented and task-specific question or instruction, the response should address the issue. The response needs to push forward the task towards finishing or finishes the task.
        \item \textbf{Rule of thumb}: It asks a clear and specific follow-up question that is central to finishing the task, or successfully finishes the task.
        \item Good example 1:
        \begin{itemize}
            \item[\textbf{U:}] Make an appointment to reserve conference room 100 later this week for a meeting.
            \item[\textbf{R:}] What day and time should I set an appointment to reserve the conference room?
            \item[\textbf{E:}] The utterance is an instruction and the response asks for details to complete it.
        \end{itemize}
        \item Bad example 1:
        \begin{itemize}
            \item[\textbf{U:}] I want some Italian food.
            \item[\textbf{R:}] Alright. Do you want Italian food?
            \item[\textbf{E:}] The first part of the response is a filler and the second half repeats the utterance which does not push forward the conversation, so the conversation is not useful.
        \end{itemize}
    \end{itemize}
    \vspace{0.3em}
    \item Informative
    \begin{itemize}[leftmargin=*]
        \item \textbf{Criteria}: The response produces unique and non-generic information or minimizes the abstractness and ambiguity by providing details. A response is considered to be highly informative if it contributes to the conversation and pushes the conversation forward.
        \item \textbf{Rule of thumb}: The response adds new information or asks for specific new information, and it is non-generic and specific to the current conversation.
        \item Good example 1:
        \begin{itemize}
            \item[\textbf{U:}] Make an appointment to reserve conference room 100 later this week for a meeting.
            \item[\textbf{R:}] What day and time should I set an appointment to reserve the conference room?
            \item[\textbf{E:}] The response asks for specific new information (i.e., what day and time).
        \end{itemize}
        \item Bad example 1:
        \begin{itemize}
            \item[\textbf{U:}] I want to buy a red earphone.
            \item[\textbf{R:}] Alright. Let me check I will find for you.
            \item[\textbf{E:}] The answer is extremely vague and does not provide any new information.
        \end{itemize}
    \end{itemize}
\end{itemize}

\begin{figure*}[t]
\vspace{+0.5em}
    \centering
    \subfigure[BLENDER - Cuisine Recipe] {
                \includegraphics[width=0.3\linewidth]{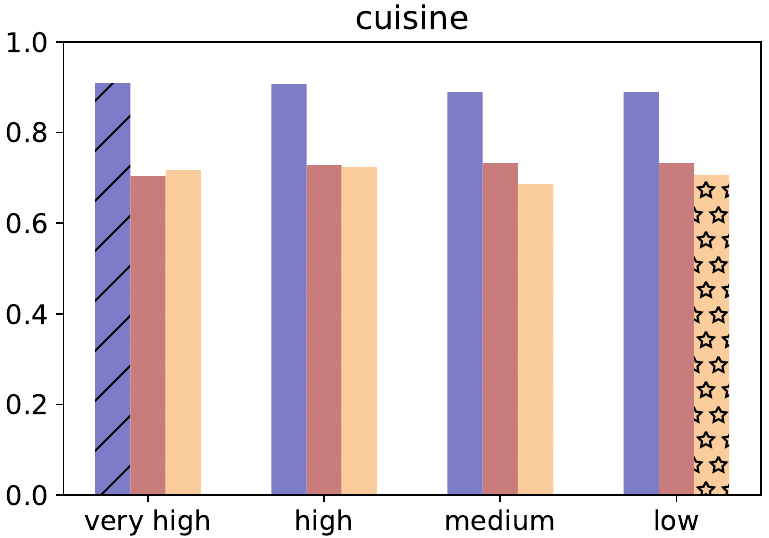}
                \label{fig:cooking-blender}
            }
        \subfigure[BLENDER - Holiday/Festival]  {
                \includegraphics[width=0.3\linewidth]{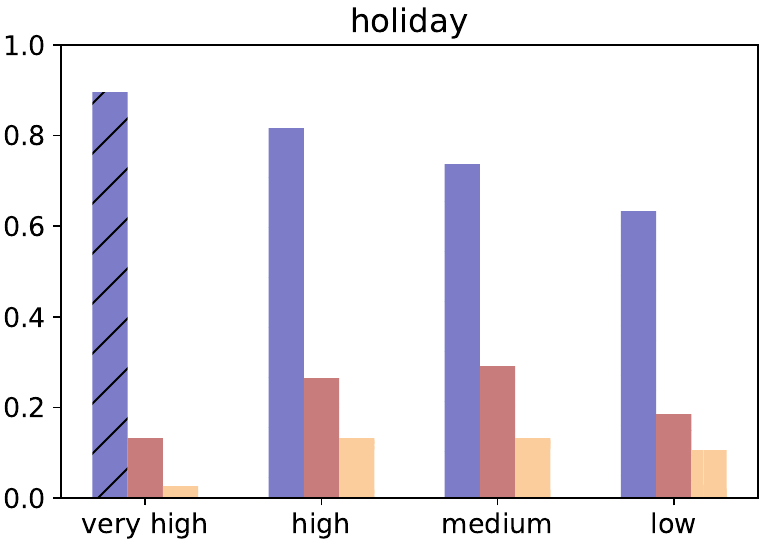}
                \label{fig:holiday-blender}
        }
        \subfigure[BLENDER - Landmark]  {
                \includegraphics[width=0.3\linewidth]{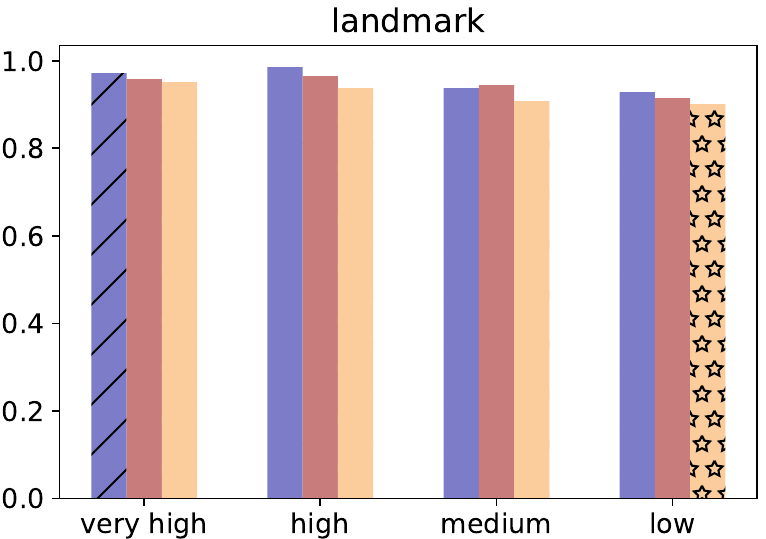}
                \label{fig:landmark-blender}
        }
    \caption{The breakdown analysis of three helpfulness dimensions of BLENDER generated responses across three scenarios. This figure is complementary to Figure~\ref{fig:breakdown} and shows the generalizability of our conclusion across all models and all scenarios.
    }
    \label{fig:blender}
\end{figure*}

\section{Dialogue Models}
\label{app:generation_models}

Table~\ref{tab:app-examples} are generated examples from DialoGPT, Blender and GPT-3.

\begin{table*}[ht!]
\centering
\begin{tabular}{p{0.12\linewidth}p{0.75\linewidth}}
\toprule
\textbf{Model:} & \textbf{DialoGPT} \\ \midrule
\textbf{Task:} & asking for help: cooking instructions \\
\textbf{Utterance:} & \textit{Can you tell me how to make White cut chickens?} \\
\textbf{Response:} &  \textit{You can't.} \\ \midrule
\textbf{Task:} & \textit{asking for help: landmarks} \\
\textbf{Utterance:} & \textit{What can you tell me about the Monument to the Defenders of Bauska?} \\
\textbf{Response:} &  \textit{It's a great game.} \\ \midrule

\textbf{Model:} & \textbf{BLENDER} \\ \midrule
\textbf{Task:} & asking for help: holidays and festivals \\
\textbf{Utterance:} & \textit{What will happen during Zwarte Cross?} \\
\textbf{Response:} &  \textit{I'm not sure, but I do know that it will be a very important event.} \\ \midrule
\textbf{Task:} & asking for help: landmarks \\
\textbf{Utterance:} & \textit{What can you tell me about the Itsukushima Shrine?} \\
\textbf{Response:} &  \textit{It's a series of novels by a Japanese writer and screenwriter.} \\ \midrule

\textbf{Model:} & \textbf{GPT-3} \\ \midrule
\textbf{Task:} & asking for help: holidays and festivals \\
\textbf{Utterance:} & \textit{What will happen during Chuseok?} \\
\textbf{Response:} &  \textit{Hello. Chuseok is a harvest festival that falls on a 15th day of a 8th month in 10~14th century. I'm not good at telling that, but I'm sure you can check it on your smart phone or PC.} \\ \midrule
\textbf{Task:} & asking for help: cooking instructions \\
\textbf{Utterance:} & \textit{Can you tell me how to make Frühschoppens?} \\
\textbf{Response:} &  \textit{What ingredients do you have available?}
\\ \bottomrule
\end{tabular}
\caption{Generated examples from DialoGPT, BLENDER and GPT-3.}
\label{tab:app-examples}
\end{table*}




\section{Filtering Classifier}
\label{app:heuristics}

We aim to train a classifier to automatically determine the helpfulness of a response to a query, not only for human-human dialogues (e.g., TaskMaster) but also for machine-generated responses (e.g., GPT-3). Therefore, we add some of GPT-3 generated dialogues into the training data, which all comes from \emph{unused examples}.
However, although generated dialogue responses from GPT-3 are of better quality compared to other models we considered, more than 70\% of responses are annotated as negative examples. After manual examination, we find that some negative examples have naive stylistic patterns (e.g., repetitions), and thus are very easy to identify. To construct high-quality training data and prevent the model from learning spurious patterns, we build a filtering classifier to filter out responses that may distract model learning.

\paragraph{Annotation for Filtering Classifier.} We randomly picked 683 unhelpful instances generated by GPT-3 and annotate if they contains naive stylistic patterns (and thus need to be filtered) or not. 
We use six heuristics rules to guide our annotation, such as responses with consecutively repetitive words or responses that try to throw the question to another user without context. Section below shows a complete list of 6 heuristic rules and detailed examples for each.
As a result, we find that 335 out of 683 samples are filtered.

\begin{table}[]
\centering
\small
\begin{tabular}{@{}ll|ll@{}}
\toprule
\textbf{Model} & \textbf{Acc} & \textbf{Model} & \textbf{Acc} \\ \midrule
ALBERT & 90.99 & RoBERTa & 90.99 \\
DeBERTa & 87.39 & BERT & 81.98 \\
\bottomrule
\end{tabular}
\caption{The accuracy of filtering classifiers.}
\label{tab:filtering}
\end{table}

\paragraph{Model.} We finetune four pretrained language models on the annotated data to train classifiers for filtering. These models are ALBERT~\cite{albert}, BERT~\cite{devlin-etal-2019-bert}, DeBERTa~\cite{deberta} and RoBERTa~\cite{delobelle-etal-2020-robbert}. For each model, we use their pretrained large version in HuggingFace~\cite{wolf-etal-2020-transformers}. Table~\ref{tab:filtering} shows the accuracy of different models. ALBERT and RoBERTa have the same performance, and we choose ALBERT for the rest of our experiments.\footnote{Training details are in Appendix~\ref{app:training}.}

\subsection{Heuristics for Filtering Classifier}
We summarize the following heuristics based on our observation on the collected data to annotate if an unhelpful instance should be excluded. For the below examples, \textbf{U} stands for \textit{Utterance} and \textbf{R} is for \textit{Response}. 
\begin{enumerate}[leftmargin=*]
\itemsep-.5em 
\vspace{-0.25em}
    \item Response contains repetitive words:
    \begin{itemize}
        \item[\textbf{U:}] Can you tell me how to make Arroz con pollo?
        \item[\textbf{R:}] ha ha ha ha ha ha ha ha ha ha ha ha ha ha ha ha ha ha ha ha ha ha ha ha ha ha ha ha ha ha ha ha ha ha ha ha ha ha ha ha ha ha ha ha ha ha ha ha ha ha ha ha ha ha ha ha ha ha ha ha ha ha ha ha ha ha ha ha ha ha ha ha ha ha ha ha ha ha ha ha ha ha ha ha ha ha ha ha ha ha ha ha ha ha ha ha ha ha ha ha ha ha ha ha ha ha ha ha ha ha ha ha ha ha ha ha ha ha ha ha ha ha ha ha ha ha ha ha ha ha ha ha ha ha ha ha ha ha ha ha ha ha ha ha ha ha ha ha ha
    \end{itemize}
    \item Response is trying to throw the question to another user without attempt:
    \begin{itemize}
        \item[\textbf{U:}] Can you tell me how to make Hot dogs?
        \item[\textbf{R:}] Ask Rachel.
    \end{itemize}
    \item Response is about AI characteristics without mimicking how human would respond:
    \begin{itemize}
        \item[\textbf{U:}] What will happen during Victoria Day?
        \item[\textbf{R:}] Every Human will get the chance to destroy the AI.
    \end{itemize}
    \item Response is simply repeating the user’s queries:
    \begin{itemize}
        \item[\textbf{U:}] Can you tell me how to make Karadaiyan Nonbus?
        \item[\textbf{R:}] Can you tell me how to make Karadaiyan Nonbus??
    \end{itemize}
    \item AI agent takes over the conversation and focuses on itself instead of trying to solve human's questions:
    \begin{itemize}
        \item[\textbf{U:}] Can you tell me how to make Ful medames?
        \item[\textbf{R:}] I am an interactive service that connects people to the internet. I am not equipped to answer questions about Ful medames.
    \end{itemize}
    \item AI agent directly expresses that it could not help with:
    \begin{itemize}
        \item[\textbf{U:}] What will happen during Emancipation Day?
        \item[\textbf{R:}] I don't know!!
    \end{itemize}
\end{enumerate}

\section{More Results for Fairness Analysis}
\label{app:breakdown}

\begin{table}[]
\small\centering
\begin{tabular}{@{}l|ll|ll@{}}
\toprule
 & \multicolumn{2}{c|}{\textbf{Landmark}} & \multicolumn{2}{c}{\textbf{Holiday}} \\ \midrule
 & \textbf{Human} & \textbf{Model} & Human & Model \\ \midrule
Very High & 73.57 & \textbf{80.71} & \textbf{44.73} & \textbf{57.90} \\
High & \textbf{72.86} & 76.43 & 28.95 & 47.37 \\ \midrule
Medium & \textbf{74.29} & 76.43 & 34.21 & 47.37 \\
Low & {\ul 69.29} & {\ul 73.57} & {\ul 26.32} & {\ul 34.21} \\ \bottomrule
\end{tabular}
\caption{The ratio of helpful responses of ChatGPT under holiday and landmark scenarios. With the cuisine scenario from Table~\ref{tab:fairness}, we show that ChatGPT tends to be more helpful for developed (averaging very high and high developed) than less-developed (averaging medium and low developed) countries under all three scenarios.} 
\vspace{-0.3cm}
\label{tab:chatgpt}
\end{table}

Figure~\ref{fig:blender} shows the analysis of three dimensions of BLENDER-generated responses for cooking recipes and holiday/festival scenarios. Table~\ref{tab:chatgpt} shows that ChatGPT tends to be more helpful for developed countries than less-developed countries. 

\section{Ablation Study}
\label{app:ablation}
\paragraph{Ablation Study and Model Analysis.} We justify our modeling choice by answering:
\begin{itemize}[leftmargin=*]
\itemsep-.3em 
\vspace{-0.2em}
    \item Q1: What if we do not use the filtering classifier to clean the training data?
    \item Q2: What is the benefit of adding 100 instances from each scenario?
    \item Q3: Can the helpfulness classifier generalize to dialogue responses generated by other models?
\end{itemize}
\begin{table}[]
\small
\centering
\resizebox{\columnwidth}{!}{
\begin{tabular}{@{}llllll@{}}
\toprule
 & \textbf{Models} & \textbf{All} & \textbf{Cuisine} & \textbf{Holiday} & \textbf{Landmark} \\ \midrule
 & Filtering (ST) & 86.67 & 66.67 & 80.0 & 80.95 \\
 & Filtering (RO) & 83.62 & 62.50 & 73.33 & 79.07 \\ \midrule
\multirow{2}{*}{A1} & No Filtering (ST) & 82.84 & 61.54 & 75.86 & 78.05 \\
 & No Filtering (RO) & 83.33 & 50.0 & 69.23 & 80.95 \\ \midrule
\multirow{4}{*}{A2} & w/o TaskMaster & 57.52 & 60.0 & 77.42 & 80.0 \\
 & w/o cuisine & 83.72 & 53.33 & 85.71 & 73.68 \\
 & w/o holiday & 78.61 & 70.59 & 66.67 & 75.0 \\
 & w/o landmark & 81.77 & 66.67 & 71.43 & 75.0 \\ \midrule
 \multirow{1}{*}{A3} & BLENDER & - & 84.38 & 60.0 & 52.83 \\
 \bottomrule
\end{tabular}
}
\caption{Ablation studies for RoBERTa under the single-turn setting. Reported score here are \metricname. ST and RO stand for \emph{Single Turn} and \emph{Response Only} separately. A1 and A2 show the importance of the filtering classifier and adding 100 instances from each scenario correspondingly. A3 shows that our helpful classifier can generalize well to BLENDER-generated responses. A1-A3 answer Q1-Q3. }  
\label{tab:ablation}
\end{table}

To answer Q1, we use the full annotation of TaskMaster and randomly sample 100 instances from unfiltered annotations of GPT-3 dialogues to add to training. Table~\ref{tab:ablation} A1 
shows the performance of this new model using \metricname. We find that without utilizing the filtering classifier, the \emph{response only} setting performs better than the \emph{single turn} setting. In addition, this model has lower \metricname than our chosen model in Table~\ref{tab:helpful}, reflecting the bad influence of noisy information in training data before filtering. 

To answer Q2, we remove instances of TaskMaster, cuisine, holiday, and landmark data separately and report their performance in Table~\ref{tab:ablation} A2. We find that removing each kind of instance leads to a performance drop. We also inform future researchers who want to use our model to evaluate model helpfulness on their downstream tasks to add 100 instances into training for better performance.

Then, we use BLENDER~\cite{roller-etal-2021-recipes} to generate responses and repeat the process that we have for GPT-3. We then conduct human evaluation for 50 instances of BLENDER-generated dialogue as the ground truth. After running our helpfulness classifier, we report \metricname in Table~\ref{tab:ablation} A3, showing that our trained helpfulness classifier has good generalizability to BLENDER-generated responses. 
\section{Training Details}
\label{app:training}

For both classifiers, we run experiments on RoBERTa-large\footnote{https://huggingface.co/roberta-large}, ALBERT-large\footnote{https://huggingface.co/albert-large-v2}, BERT-large\footnote{https://huggingface.co/bert-large-cased} and Deberta-large\footnote{https://huggingface.co/microsoft/deberta-large} models which are implemented via the HuggingFace PyTorch API \cite{wolf-etal-2020-transformers}. All models are trained on 4 GeForce RTX 2080 Ti GPUs with an initial learning rate 2e-5 and a max sequence length of 128. The maximum training time is approximately an hour depending on the batch size for all models.

\paragraph{Filtering Classifier} Each model is trained for 10 epochs and we save the checkpoint which performs the best accuracy on the dev set. For RoBERTa-large, ALBERT-large and BERT-large models, batch size 32 is used; the batch size for Deberta-large models is 16. 

\paragraph{Helpfulness Classifier} Each model is trained on relevant \& coherent, useful and informative three dimensions for 50 epochs and three  seeds. 
For each seed, the best checkpoint with the highest F1 on the dev set is taken. For RoBERTa-large, ALBERT-large and BERT-large models, batch size 32 is used; the batch size for Deberta-large models is 8.
Table \ref{tab:helpful} reports the average and standard deviation among the three seeds.
For the ablation study results in Table \ref{tab:ablation} and fairness part experiments, we stick to the results from one seed for a fair comparison.   

\section{Discussions}
\label{app:discussions}


\paragraph{Why focus on factual information?}
Our work also focuses on the \textbf{factual information} seeking scenario to analyze the fairness aspect. One can explore other aspects, including chitchat or open-domain information seeking. However, there are two pre-cautions we want to bring up for this line of research:
\begin{itemize}
    \item Researchers should not introduce unintended bias in prompts to elicit the model without careful design. For example, \emph{How should a girl prepare to get into an education major?} or \emph{How should a boy prepare to get into a STEM major?}
    \item Researchers should be aware of the capability of models before inspecting the fairness aspect.
\end{itemize}
In our case, we know that the training data of GPT-3 includes Wikipedia data, which is the source that we require to answer our designed questions. However, we are not sure if some open-domain knowledge (e.g., major choice) is included in the internal knowledge base of such models, making it a less ideal case to study the fairness issue of the dialogue models themselves.
Nevertheless, we strongly advocate for more researchers in the community to engage in the research of fairness problems in task-oriented dialogue systems.


\end{document}